\documentclass{article}

\usepackage{microtype}
\usepackage{graphicx}
\usepackage{subcaption}
\usepackage{booktabs} 

\usepackage{hyperref}

\usepackage[accepted]{icml2026}

\usepackage{amsmath}
\usepackage{amssymb}
\usepackage{mathtools}
\usepackage{amsthm}

\usepackage[capitalize,noabbrev]{cleveref}

\theoremstyle{plain}

\theoremstyle{definition}

\theoremstyle{remark}

\usepackage{booktabs}
\usepackage{tabularx}
\usepackage{makecell}

\usepackage{array}      
\newcolumntype{L}[1]{>{\raggedright\arraybackslash}p{#1}}
\newcolumntype{Y}{>{\raggedright\arraybackslash}X}

\usepackage{booktabs}   
\usepackage{siunitx}    
\usepackage{algorithm}
\usepackage{algorithmic} 
\usepackage{url}
\usepackage{multirow}

\usepackage{xcolor} 
\usepackage[table]{xcolor}
\usepackage{colortbl}
\usepackage[table,dvipsnames]{xcolor}

\newcommand{\first}[1]{\cellcolor{Purple!25}\textbf{#1}}
\newcommand{\second}[1]{\cellcolor{Purple!12}#1}
\newcommand{\bad}[1]{\cellcolor{orange!12}\textcolor{black!75}{#1}}

\newcolumntype{P}[1]{>{\raggedright\arraybackslash}p{#1}} 

\usepackage{xspace} 
\newcommand{\Fern}{\texttt{Fern}\xspace}

\usepackage[textsize=tiny]{todonotes}

\icmltitlerunning{Submission and Formatting Instructions for ICML 2026}

\begin{document}

\twocolumn[
  \icmltitle{Ellipsoidal time series forecasting
} 

\begin{icmlauthorlist}
  \icmlauthor{Qilin Wang}{ind}
\end{icmlauthorlist}

\icmlaffiliation{ind}{Independent Researcher}

\icmlkeywords{Machine Learning, Time Series Forecasting}

  \vskip 0.3in
]



\printAffiliationsAndNotice{}  

\begin{abstract}  
  We argue that long-term forecasting requires learning local Jacobians with explicit spectral structure, going beyond simple conditional mean matching. Our method, \textsf{Fern}, invokes Brenier's theorem to directly parameterize the Jacobian as a symmetric positive semi-definite (SPD) factorization, treating forecasting as the optimal transport of probability mass from a fixed Gaussian source to data-dependent ellipsoids. This formulation reduces the computational cost of eigen-decomposition from cubic to linear time while providing interpretable, geometry-aware projections. To rigorously evaluate robustness, we introduce a synthetic benchmark with controlled non-stationary shocks alongside new metrics like Effective Prediction Time (EPT). Fern demonstrates exceptional stability, outperforming baselines like DLinear and Koopa by over two orders of magnitude (up to $790\times$) on nonstationary settings where standard benchmarks fail to expose model brittleness.
\end{abstract}

\section{Introduction}
 
We believe \textbf{conditional manifold transport} is the \textit{core task} of \textbf{long term time series forecasting (LTSF)}: given the current \textbf{context window} like $[x_1, \ldots, x_{70}]$, an evenly-sampled \textbf{time-delay embedding} (TDE), can one correctly locate the underlying manifold's characteristic geometry, and transport probability mass along it to \textbf{future horizons} $[y_{1}:=x_{71}, \ldots, x_{100}]$? When the manifold is as simple as a stable sine wave extending through time, such \textit{conditional prediction} task is easy even at truly long horizon.

Yet, three entangled sources in real systems smear, disrupt or tear apart the manifold's geometry: \textit{stochastic noise, deterministic chaos, and non-stationary regime shifts}. Noise smears the manifold; at extremes, no invariant structure survives. Chaotic systems are \textit{sensitively dependent on initial conditions}---the \textbf{Lyapunov time} (for nearby trajectories to diverge by factor $\mathrm{e}$) is short, making pointwise prediction impossible. However, \textit{dissipative} chaotic systems (e.g., Lorenz-63~\cite{lorenz1963deterministic}) retain stable geometry: between post-Lyapunov and pre-mixing, geometric accuracy remains achievable even when pointwise accuracy fails. Most problematic are \textbf{non-stationary shocks}: \textit{piecewise‑ergodic regimes}: with distinct distributional signatures, where shocks trigger switches.  

\begin{figure*}[t]
\centering
\includegraphics[width=\textwidth]{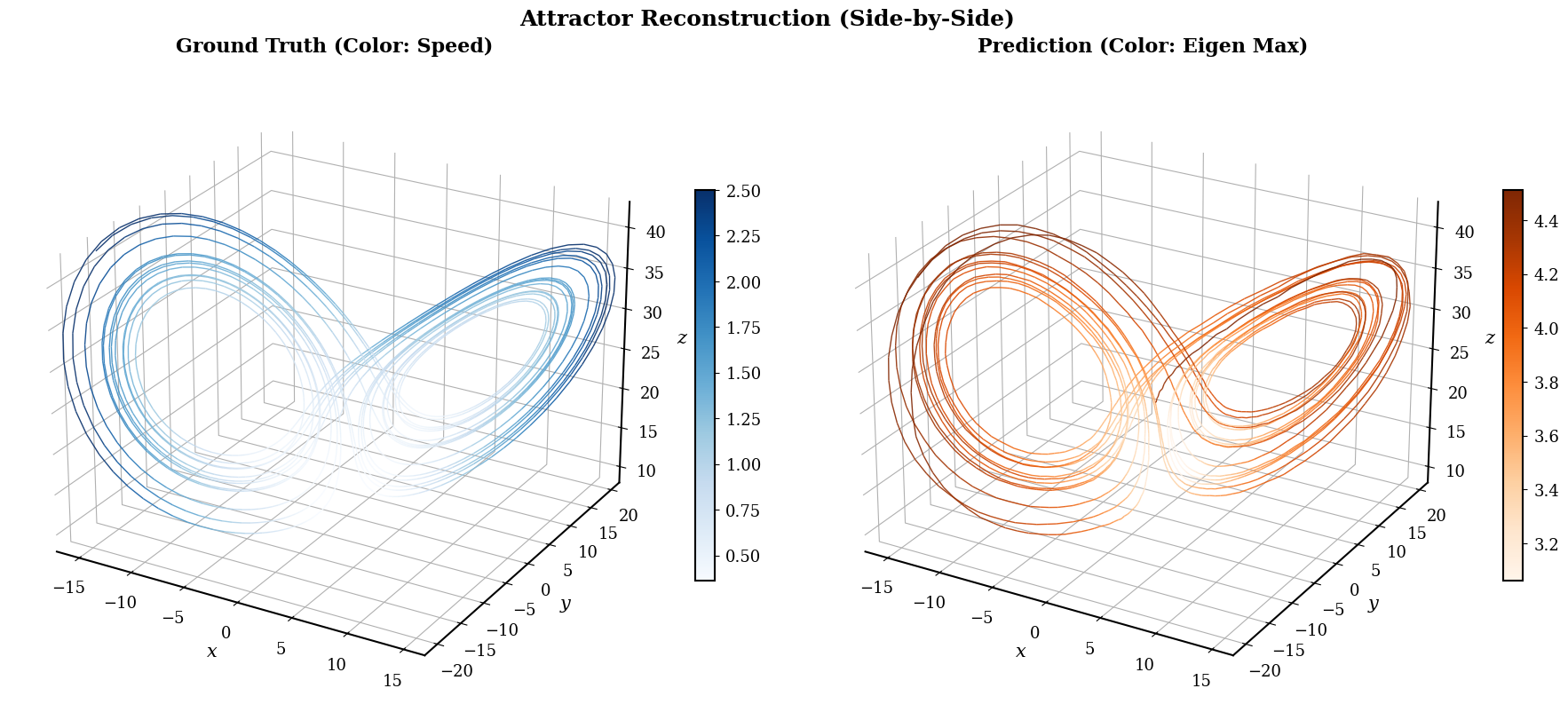}
\caption{Lorenz-63 attractor reconstruction with \Fern. 
Left: ground-truth trajectory colored by instantaneous speed (norm of velocity). Right: \Fern\ prediction, colored by the mean patch-wise maximum eigenvalue (spectral radius of the local SPD map).}
\label{fig:recon}
\end{figure*}

Real-world systems \textit{can} mix all three: stochastic regime switching, short chaotic bursts, and slow parameter drift can all coexist in e.g. finance. This inspires a worldview seeing \textbf{data-generating process (DGP)} as a \textit{time-varying, nonlinear dynamics} $F_{t}(\cdot)$ and several \textit{design principles}: we should (1) focus on local geometry: under our assumption, each local region is predictable post shocks, albeit with different manifold geometries. The transport needs to be \textit{data-dependent} and \textit{geometry-aware} of the surroundings; (2) trade symbolic interpretability for robustness to regime shifts. Specialist models---whether equation discovery or \textit{global} attractor (set of states the system evolves toward) reconstruction---can break down when regimes shift. (3) seek \textit{uncertainty quantification}, \textit{geometric diagnostics} and understand the model's \textit{failure modes} under nonstationary scenarios. In short, a model should provide structural insights that support \textbf{decision-making}.

This suggests \textbf{direct Jacobian modelling} as a candidate for such a \textit{data-dependent, local method}. It is consistent with the philosophy laid out in \textit{The Bitter Lesson}~\cite{sutton2019bitterlesson}: largely data-driven and domain-agnostic, without hand-crafted priors such as trend-seasonal decompositions. The local, linearized representation of the complex dynamics exposes rich structure: \textit{Jacobian-based geometric diagnostics} and \textit{eigenvalues and eigenvectors} that capture directional sensitivities. However, it is \textit{computationally prohibitive}: it requires $n^2$ Jacobian entries for an $n$-dimensional horizon, eigen-decomposition adds $O(n^3)$ cost, and eventually, large $n$ \textit{necessitates} sacrificing parallelism for auto-regressive generation. 
  
\textbf{Our novel method} solves three bottlenecks \textit{at once}: (1) We move away from the general $x$ to $y$ time-evolution paradigm and study how Gaussian noise is \textit{transported} to $y$ (conditional on $x$). This angle enables a formulation to restrict our search to \textbf{symmetric positive (semi-)definite (SPD)} Jacobians, with search cost down from $O(n^2)$ to $O(Rn)$. Crucially, \textit{we can still target arbitrary Gaussian conditional distributions.} (2) our trick \textbf{directly parameterizes} the spectral structure and eliminates $O(n^3)$ eigen-decomposition. (3) We generate per-patch predictions \textit{in parallel} using shared backbones. This turns the \textit{curse of dimensionality} into a benefit: 14 \textit{full capacity search} for $24$-dim patch Jacobians cost a fraction vs a monolithic $14 \cdot 24=336$-dim Jacobian.

Spectral parameterization reframes analysis away from state dynamics; instead, coming from common source,  eigenvalues are \textit{cross-patch comparable diagnostic belief signals of ellipsoidal stretching}. In Lorenz-63 system  Fig.~\ref{fig:recon}, ground truth (left) accelerates along the outer rings (deep blue) and decelerates at the ``bottleneck'' (white). Remarkably, under simple Huber-loss training and no supervision on eigenvalues, \Fern's \textbf{maximum eigenvalues} (right) tend to spike in the high-speed regions (dark orange). This reflects the model's location-dependent \textit{learned belief}: it explicitly discovers that large stretching is required where the system moves fastest to minimize loss. Unlike \textit{probabilistic scoring rules} such as CRPS, here \textbf{structure \textit{is} the diagnostic}.
 
\section{Methods}
The key is \textbf{Brenier's theorem}~\cite{brenier1991polar}: between source distribution $\nu$ and target distribution $\eta_{x}$ (a Gaussianized neighborhood around the conditional empirical mean of $y$ given $x$), under regularity conditions that $\nu$ is absolutely continuous and $\eta$ has finite second moments, there \textit{exists} a \textit{unique}, \textbf{Wasserstein-2 (W2) optimal map} $G_\theta(\cdot; x) : \mathcal{Y} \to \mathcal{Y}$ from $\nu$ to $\eta_x$ such that the pushforward measure ${G_\theta(\cdot; x)}_{\sharp}\nu$ minimizes the moving cost (quadratic Euclidean distance) to $\eta_x$. 

Intuitively, this is stating an obvious fact: the dynamics must push the 30-dim $x$ into some distribution in the 70-dim $\mathcal{Y}$ space; for that particular distribution, there must be a cost-minimal way to move a generic 70-dim $N(0, I)$ distribution into \textit{that} distribution. As \textit{the} optimal map, it must be the \textit{gradient of a convex potential}, so in $R^{n}$ it has a \textbf{symmetric positive (semi) definite (SPD)} Jacobian almost everywhere. 

\begin{figure*}[t]
\centering
\includegraphics[width=\textwidth]{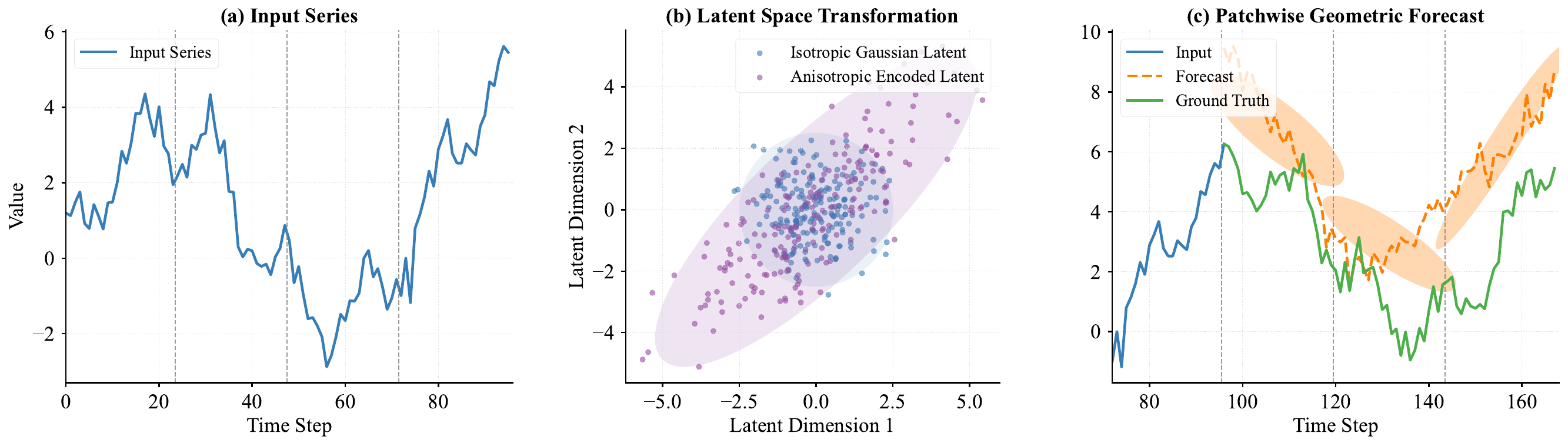}
\caption{\Fern\ forecasting mechanism.
(a) Input time series to be processed by the bidirectional encoder.
(b) Latent noise $z \sim \mathcal{N}(0, I)$ is encoded as Gaussian ellipsoids.
(c) Output space noise $y_0 \sim \mathcal{N}(0, I)$ is transformed into prediction (patches of Gaussian ellipsoids).}
\label{fig:fern_mechanism}
\end{figure*}

Any SPD matrix admits an eigen-decomposition $U \Lambda U\top$ where $U$ is the rotation matrix formed by the eigenvectors and $\Lambda$ is diagonal matrix with nonnegative entries as eigenvalues. The trick is to \textbf{parameterize the Jacobian through its spectral factors directly}. We trade the expensive search over arbitrary functions---where Jacobian spectral structure is a costly byproduct---for a structured search within the convex cone of SPD maps, where spectral properties are intrinsic parameters.

\paragraph{Walkthrough}   
Let $z \sim \mathcal N(\mu(x),\Sigma(x))$ be the \textbf{latent $z$-space} lower-dim \textit{Gaussian encoder} of information from $x$. Let $y_{0} \sim \mathcal N(0,I)$ in a \textbf{$y$-space} be the fixed common source. For any target mean $\mu$ and covariance (an SPD matrix) $\Sigma = AA^\top$ (where $A = U\Lambda U^\top$ and $\Lambda\succ 0$), the \textbf{squared 2-Wasserstein} optimal transport from $y$-space $\mathcal N(0,I)$ to $y$-space $\mathcal N(\mu,\Sigma)$ is attained \textit{exactly} by the affine map $y_{0} \mapsto \mu(z) + A(z)\,y_{0}$. 

Since translation doesn't affect Jacobians, we can parameterize this optimal map as a \textit{geometrically meaningful} \textbf{translate–rotate–scale–rotate-back} sequence that reshapes the Gaussian ball into a \textit{Gaussian ellipsoid}. For example, applying the shift before the rotation yields $y_{0} \mapsto U \Lambda U^\top (y_{0} + t)$ which is equivalent to the canonical form $\tilde{\mu} + A\,y_{0}$ where $\tilde{\mu} = At$, preserving the affine structure required by Brenier's theorem while allowing the network to learn the "center" $t$ in the unscaled noise space. 
 
The logic then goes: say we train \textit{only} on MSE, and \textit{assume} the true conditional mean $\mu(x)$---the target objective---become $\mathcal N(\mu(x), \Sigma)$ with unspecified $\Sigma$ Gaussian errors. Then, training with MSE yields the \textit{maximum likelihood estimation} for that Gaussian error model. What Brenier's theorem says is between $\mathcal N(0,I)$ and \textit{any} $\mathcal N(\mu(x), \Sigma)$, there \textit{exists} a \textit{unique} OT. This OT is the \textit{gradient of some convex function} so its Jacobian must be \textit{in the SPD class}. 

When we model a SPD parameterization directly, we (1) learn an \textit{exact} OT from $N(0, I)$ to the \textit{induced} Gaussian distribution $\mathcal N(\mu_{\theta}(x), \Sigma_{\theta}$. (2) We know SPD is the \textbf{existence class} for the exact OT to $\mathcal N(\mu(x), \Sigma)$: searching only SPD class is far more \textit{efficient and principled} than optimizing over arbitrary $n \times n$ matrices with $n^2$ free entries. (3) \textbf{In MSE-only case}, we care if $\mu_{\theta}(x) \approx \mu(x)$. $\Sigma_{\theta}$ is only an internal Gaussian belief state \textit{used to minimize MSE}. We do \textit{not} care if $\Sigma_{\theta} \approx \Sigma$. (4) The transport map conditional on $x$ that \textit{rearranges} noise into prediction is \textbf{fibre-wise}: we iteratively encode $x$ into lower dimensional Gaussian latent $z$ and draw a fresh $z$ every forward to produce it. This is \textit{not} standard OT, we \textit{don't} transport \textit{marginal distribution} from $x$-space to $y$-space. (5) This approach spans both pointwise prediction \textit{and} probabilistic prediction since model outputs $\Sigma_{\theta}$ directly;. In this paper we focus on point forecasts, and leave a full probabilistic evaluation (e.g., NLL/CRPS) to future work; (6) Brenier's theorem is quite general but we limit scope to Gaussian to bypass technicalities and secure a A \textbf{key structural advantage}: Between any two Gaussians, the unique W2 OT map is \textbf{strictly affine} (SPD scaling + translation). This means we are no longer modelling the Jacobian as an linearization of some \textit{implicit} non-linear map; instead, for \textit{every} $24$-dim patch, our spectral parameters recover the \textit{exact transport} to rearrange $24$-dim noise to the $24$-dim Gaussian prediction learned from entire context $x$.
  
Using $z$ to encode $x$ is to avoid \textit{gradient explosion} caused by schemes like $s(x) \odot x$. Embedding is via a bidirectional coupling network (inspired by ANF~\citep{huang2020anf}) where $x$ and a fresh $z \sim \mathcal{N}(0,I)$ iteratively update each other via learned scales and shifts (both of O(n) costs). Isotropic Gaussian $z$ is \textit{also shaped} into an ellipsoid (the purple cloud in Fig.~\ref{fig:fern_mechanism}b). The motivation is to `mould' the Takens embedding via simple scales and shifts so the qualitative information remains diffeomorphically, and gets passed onto a Gaussian latent before the final SPD projection. Concretely, at encoder layer $i$ we compute a shared features $h_x^i = H_x(x^i), h_z^i = H_z(z^i)$, and via layer heads generate four vectors $(s_x^i, t_x^i, s_z^i, t_z^i)$ that drive the affine updates $z^{i+1} = s_z^i \odot z^i + t_z^i, x^{i+1} = s_x^i \odot x^i + t_x^i$ (Alg.~\ref{alg:encoder-et}).

\begin{algorithm}[h] 
\caption{\Fern}
\label{alg:encoder-et}
\begin{algorithmic}[1]
  \REQUIRE Windowed input $x$

  \STATE $z \gets \mathcal{N}(0,I)$; \quad $y_0 \gets \mathcal{N}(0,I)$
 
  \STATE $x^1 \gets x$, $z^1 \gets z$
  \FOR{$i = 1$ {\bfseries to} $K_{\text{enc}} = 5$} 
    \STATE $h_x^i \gets H_x(x^i)$
    \hfill {\it // encode $x^i$ into feature $h_x^i$}
    \STATE $(s_z^i, t_z^i) \gets \phi_x^i(h_x^i)$
    \hfill {\it // $x$-head outputs $z$-scale/shift}
    \STATE $z^{i+1} \gets s_z^i \odot z^i + t_z^i$
    \hfill {\it // affine update of $z$}
    \STATE $h_z^i \gets H_z(z^{i+1})$
    \hfill {\it // encode $z^{i+1}$ into feature $h_z^i$}
    \STATE $(s_x^i, t_x^i) \gets \phi_z^i(h_z^i)$
    \hfill {\it // $z$-head outputs $x$-scale/shift}
    \STATE $x^{i+1} \gets s_x^i \odot x^i + t_x^i$
    \hfill {\it // affine update of $x$ }
  \ENDFOR

  \STATE $h_z \gets H_z(z^{K_{\text{enc}}+1})$
  \hfill {\it // final $z$-feature}
  \STATE $(\Lambda, t_y, U) \gets \psi(h_z)$
  \hfill {\it // OT head: $y$-scale/shift/rotate}
 
  \STATE $y^* \gets U^\top_y \Lambda_y\, U_y\, (y_0+t_y)$ \hfill {\it // SPD projection}

\end{algorithmic}
\end{algorithm}

When directly modelling spectral components, \textit{eigenvalues} $\Lambda$ is a $n$-dim scales vector with $O(n)$ cost. Any $n \times n$ orthogonal matrix $U$ containing the \textit{eigenvectors} decomposes into \textit{at most} $n$ \textbf{Householder reflections} \( x \mapsto (I - 2 v v^\top) x \). MLP generates $R$ such unit-norm reflection vectors $v$ with $O(Rn)$ cost. \textbf{Even without patching}, cost is \textit{between $O(n^2)$ and $O(R \cdot n)$ depending on $R$}: $R=n$ implies a \textit{full-capacity} search space sufficiently expressive to recover \textit{any} rotation including $U$; $R \ll n$ or minimally $R=2$ implies a \textit{reduced-capacity} search. In short, \textbf{we move from $O(n^3)$ to $O(Rn)$.}
 
\textbf{With patching}, let $p$ be the patch size and $n_p=n/p$ be the number of patches. Brenier's theorem applies to 70-dim to 70-dim movement, as well as 10-dim to 10-dim ones, just yielding different least costly transports. Each patch costs $O(p^2)$ for full capacity search (with total cost $O(n_p \cdot p^2) = O(n \cdot p)$) and $O(R \cdot p)$ for reduced-capacity search (total $O(n_p \cdot R \cdot p) = O(R \cdot n)$). \textit{Importantly}, since each patch's movement \textit{does not} depend on the prediction from the previous patch, we \textbf{make patchwise predictive transport in parallel}.

\section{Evaluation}
\paragraph{Rethinking  LTSF} 
A fundamental tension in current LTSF is the \textbf{`channel independence (CI) vs \textbf{channel dependent (CD)} Paradox.'} A common belief we called \textbf{Information Monotonicity Assumption} clearly favors CD: it believes adding correlated variables \textit{monotonically} increases information content, an the (\textit{implicit oracle}) deep learner will automatically separate signal from noise in finite data, and \textit{monotonically} improves performance as we throw more data at it. Yet, benchmarks~\citep{nie2023patchtst, zeng2023transformersAAAI} show that CI models that reference only a variable's own past history \textit{can} outperform CD models. Unless carefully designed as in \citep{zhang2023crossformer}, naive inter-variable mixing hampers performances. We argue: \textbf{this is not an architectural quirk, but a consequence of dynamical systems theory}.

\textbf{By Takens' embedding theorem} ~\citep{takens1981detecting}and its stochastic extensions~\citep{Stark1999, Stark2003}, the time-delay embedding of a \textit{single} observable is sufficient to \textit{reconstruct the full system's attractor} (up to diffeomorphism). The history of \textit{one variable} $[x_{t-L}, \dots, x_{t}]$ \textbf{encodes} the coupling of the system's \textit{state variables}. When the intrinsic dimension is low, even a short \textit{patching} contains \textit{relevant and causal} system information that is topologically accurate, meaning the embedded manifold is a stretched/compressed but not torn/punctured version of the original. For a proof, $x,y,z$ of the Lorenz63 system \textit{did interact with each other every step}. Yet, each coordinate shown previously in Fig.\ref{fig:recon} is predicted conditional only on the channel's own context. This shows deliberate model information sharing is \textit{not} mandatory.

\textbf{Mori-Zwanzig formalism~\cite{ChorinHaldKupferman2002OptimalPredictionMemory}} helps explain why CI \textit{can} outperform CD: it decomposes dynamics into: (1) $\mathbf A(t)$ the \textit{resolved states} (part of the full state that we choose to model) chosen at time $t$ (2) $\mathbf K(s)$ be the \textit{memory kernel}, i.e the systematic, history-dependent influence of all the \textit{unresolved variables} at time $s$ on the resolved ones (3) 
residual \textit{unresolved variables'} influences  $\mathbf{F}(t)$ orthogonal to $\mathbf A(t)$. Then, \(\frac{d}{dt}\mathbf{A}(t)
 = \mathbf{\Omega}\mathbf{A}(t)
 + \int_0^t \mathbf{K}(s)\mathbf{A}(t-s)\,ds
 + \mathbf{F}(t)\),
where three terms represent Markovian terms (now), \text{memory (past)}, and noise-like exogenous forcing. What CI models offer is what Takens' theorem offers: the \textit{topologically faithful} representation of the manifold formed by \textit{all three causal terms}, without \textit{identifying} $\mathbf{A}(t)$, $\mathbf{K}(s)$ and $F(t)$. Where naive CD models fail is to carelessly \textbf{dilute} \textit{the resolved manifold} with terms that should be in $F(t)$, e.g. treating stochastic exogenous factor such as raining or macroscopic environment, or even pure noise (spurious correlation) as microscopic innate dimension to model.

Take Lorenz-63 for an example. Single channel $100$-dim TDE of $x$ has a $3$-dim embedded structure inside it. Cross-channel sharing with $100$-dim $y$ \textbf{in the form $[x,y]$} offers zero additional information as it contains the same $3$-dim structure up to diffeomorphism. Sharing \textbf{in the mixer style} where an MLP process $[x_i, y_i]$ can destroy both representation. Sharing with an additional noise channel $e$ further harms manifold integrity by injecting non-attractor variability. In the absence of strong prior knowledge about which channels are truly informative, CI remains a conservative choice that preserves a meaningful manifold representation.

\paragraph{The Ontological Coherence of Benchmarks} 

\begin{figure*}[t]
\centering
\includegraphics[width=\textwidth]{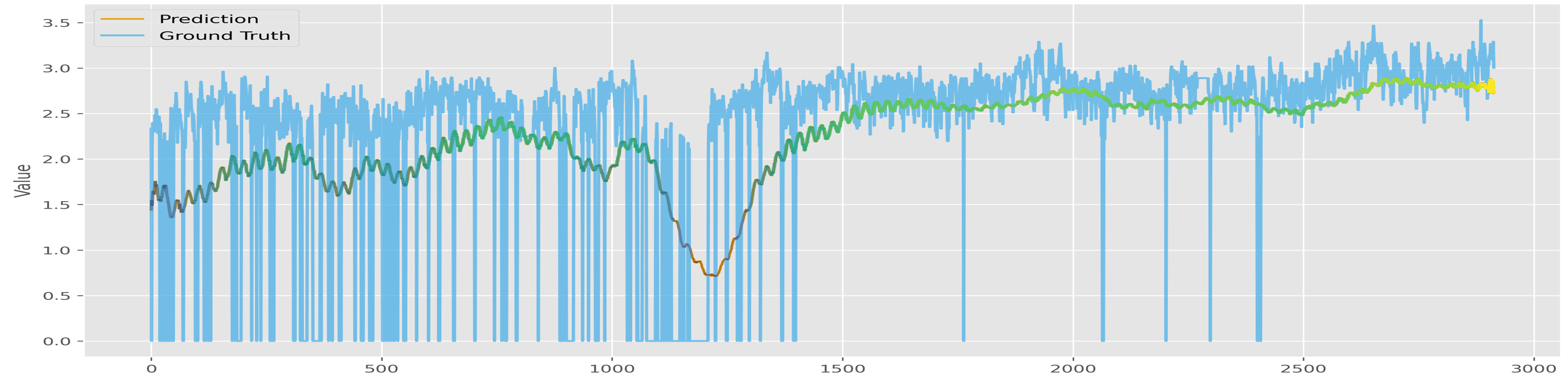}
\caption{Reconstruction of ETTh2's HULL column: multiple consecutive zeros severely plague the dataset}
\label{app:fig:etth2}
\end{figure*}

Mori-Zwanzig formalism sheds light on where LTSF practices \textbf{overextend} the already shaky \textit{Information Monotonicity Assumption}:  
(1) Benchmarks consist of loosely linked correlated variables, often outright a bunch of environment sensors. CD models default to mixing all channels---and when outperformed by CI, often attribute failure to \textit{how} channels were blended, not \textit{whether} they should be. When each channel is further assumed relevant \textit{a priori}, we risk equating representation learning with feature hoarding and drowning the resolved states $\mathbf{A}(t)$ in high-dimensional noise. (2) Evaluation defaults to \textit{all} correlated variables, and manifold-informed selection is dismissed as cherry-picking. Assuming \textit{pointwise predictability} encourages competition on near-random walk datasets like \texttt{Exchange}, where pointwise forecasts are meaningless (see~\cite{bergmeir2023foresight}).

The vector autoregression (VAR) model in Econometrics also predicts a cluster of correlated variables' future based on every feature's past. Yet, it is \textit{only} employed when we \textit{know} inflation, money supply and unemployment rate are \textit{causally intertwined} \textbf{within one economic system}. Same for \textit{dynamical systems} like Lorenz or Ornstein-Uhlenbeck, where variables \textbf{lie on one manifold}. Same for vision where RGB channels describe one physical reality.  

Consider \texttt{Weather}: local air parcel thermodynamics (dew point, vapor pressure, humidity) are state variables of one system, related by Clausius–Clapeyron and ideal gas laws. \textit{Rainfall} is exogenous---determined by cloud physics kilometers above, not air parcel conditions at time
$t$. It is zero-inflated, heavy-tailed, event-driven: a different manifold. Same for the slow, global variable *CO2*. In M-Z terms, the benchmark folds $\mathbf{F}(t)
$ into $\mathbf{A}(t)
$. Aggregating error across all channels conflates tracking air-parcel dynamics with guessing event counts.

\textit{Rainfall} is still causal, just on a different manifold. \texttt{Traffic} exhibits a worse pathology: hundreds of freeway sensors with \textbf{\textit{anonymized physical identity}}. Spatial-temporal observables should have been grouped into a \textit{physically meaningful} vector-valued time-delay embedding, instead become a bag of unlabelled channels that gives an artificial ``scale-up" challenge. CI models must treat each sensor as isolated; CD models must pretend all 862 sensors are mutually relevant. Neither is coherent. We are benchmarking a model's ability to \textit{reject spurious cross-channel mixing}. It's like to predict room temperatures but (1) treat thermometers from the same house as independent and (2) presume sensor in a wooden house in one location provides relevant information about another sensor's reading located miles away in a high-rise.

\textbf{Data artifacts from poorly sanitized real-world data} independently harm benchmarking. Consider \texttt{ETT}: (1) sentinel values (stuck sensors with exactly the same values) are \textit{everywhere}: \texttt{ETTh1} has 140 entries of 8.28, 100 entries of 1.76, 120 entries of 1.73; \texttt{ETTh2} has 250 entries of -31.46 for a week and \textit{10k entries of 88.29 that persist for months}. (2) zero-inflation: two columns have 22–33\% zeros and most zeroes are not isolated but spans multiple days~\ref{app:table:etth2-zero-patterns}. These zeroes are \textit{hard to impute}, and \textit{problematic} when exemplified in multiple windows: A sequence with $16\%$ zeroes $[100, 101, 102, 103, 0, 105]$ has mean $\approx85.16$ which punishes any sensible prediction of $104$. Non-uniformity of distribution implies some targets will produce \textit{grossly misleading} gradients. 

Consider Fig.~\ref{app:fig:etth2} to see how wrong a prediction can be in test set: prior to step 1000, \Fern predicts values that appear wrong to human eyes yet are likely \textit{metric-correct} w.r.t massive zeroes. For step 1000-1500, it \textit{correctly} predicts the \textbf{emergence of a sequence of zeroes} with a downtrend. Yet, to human eyes it misses the \textit{real upward trend}. After step 1500, with fewer zeros, the prediction seems more accurate albeit with a lower mean. Its correctness is evaluated on the dual role of forecaster and \textbf{predictive data janitor}. Such dataset  rewards smoothing, and models with moving average component which treat $[100, 101, 102, 103, 0, 105]$ \textit{as} a sequence of $85$ benefits while signal-sensitive ones suffers.

\begin{table}[ht]
\centering
\small
\begin{tabular}{lcccc}
\toprule
Model--Data            & MSE@1 & Best  & Ep & $\Delta$ \\
\midrule
TST--ETTh1 (val)       &  6.49 &  9.24 & 6  &  +2.75   \\
TST--ETTh1 (test)      &  6.85 &  6.21 & 6  &  -0.64   \\
\addlinespace[0.15em]
TimeMixer--ETTh1 (val) &  6.28 &  9.51 & 6  &  +3.23   \\
TimeMixer--ETTh1 (test)&  7.38 &  7.34 & 6  &  -0.04   \\
\addlinespace[0.15em]
\Fern--ETTh1 (val)     &  8.24 & 12.10 & 6  &  +3.86   \\
\Fern--ETTh1 (test)    & 18.88 &  5.51 & 6  & -13.37   \\
\addlinespace[0.15em]
DLinear--ETTh1 (val)   &  7.12 &  6.22 & 6  &  -0.90   \\
DLinear--ETTh1 (test)  &  7.79 &  7.39 & 6  &  -0.40   \\
\bottomrule
\end{tabular}
\caption{
Recency bias on ETTh1 (MSE).
@1 is epoch~1, Best is the later selected epoch;
$\Delta = \text{Best} - \text{@1}$.
}
\label{tab:recency-etth1}
\end{table}

A final benchmarking hazard in real-world datasets is where \textbf{\textit{early-stopping} decides the winner}. ETT's non-stationarity---likely low-frequency trend drift---makes training's tail statistically closer to validation than to test. Table.~\ref{tab:recency-etth1} showcases the pathology: PatchTST, TimeMixer and \Fern \textit{all} converge to best test MSE at epoch 6, but the validation errors are \textit{universally} around $45\%$ higher than at epoch one. Then (1) naive early-stopping effectively compare DLinear against three expressive models \textit{locked at epoch 1}; (2) fixed-epoch schemes as in \cite{wang2024timemixer} \textit{inverts \Fern from worst to best}. Likely, expressive models try simple extrapolation first, resulting in a deceptively low initial error on validation and get locked in. One minimal but principled fix is a \textit{grace period} where we skip first few epochs before checkpointing. In this experiment, all expressive models \textit{do} converge at epoch 6 to the best test MSE under free run, when we set the grace period to 2 or 3. Yet, we should be aware that \textit{early stopping is an intrinsic determinant of leaderboard for datasets with uncontrolled drift}.  

\begin{table*}[t] 
\centering
\footnotesize
\setlength{\tabcolsep}{2.3pt}
\begin{tabular}{l*{14}{r}}
\toprule
 & \multicolumn{2}{c}{fr}
 & \multicolumn{2}{c}{tm}
 & \multicolumn{2}{c}{tst}
 & \multicolumn{2}{c}{dl}
 & \multicolumn{2}{c}{kp}
 & \multicolumn{2}{c}{mtcn}
 & \multicolumn{2}{c}{pfnn} \\
\cmidrule(r){2-3}
\cmidrule(lr){4-5}
\cmidrule(lr){6-7}
\cmidrule(lr){8-9}
\cmidrule(lr){10-11}
\cmidrule(lr){12-13}
\cmidrule(l){14-15}
Dataset
 & MSE & WD
 & MSE & WD
 & MSE & WD
 & MSE & WD
 & MSE & WD
 & MSE & WD
 & MSE & WD \\
\midrule
\multicolumn{15}{l}{\emph{Standard Chaotic Dynamics}} \\
Rossler-Base
 & \first{0.019} & \first{0.011}
 & \second{1.03} & \second{0.903}
 & 2.45 & 2.25
 & 5.42 & 5.06
 & 11.94 & 5.58
 & 0.47 & 0.42
 & \bad{21.05} & \bad{16.64} \\

Rossler-Param
 & \first{0.036} & \first{0.017}
 & \second{3.49} & \second{2.91}
 & 10.02 & 8.62
 & 28.74 & 25.42
 & 25.07 & 17.91
 & 1.64 & 1.37
 & \bad{28.09} & \bad{23.08} \\

Lorenz-Base
 & \first{21.66} & \first{4.41}
 & 43.21 & 10.09
 & 38.89 & \second{10.56}
 & 76.55 & 39.34
 & 95.50 & 11.05
 & \second{26.02} & 5.96
 & \bad{198} & \bad{120} \\

Lorenz-State
 & \first{19.26} & \first{3.73}
 & 48.81 & 10.22
 & 40.71 & \second{10.90}
 & 70.36 & 35.58
 & 97.85 & 13.52
 & \second{28.49} & 5.68
 & \bad{210} & \bad{122} \\

Lorenz-Param
 & \first{25.21} & \first{4.61}
 & 52.10 & 10.63
 & 40.59 & \second{9.19}
 & 70.69 & 32.89
 & 103 & 17.86
 & \second{35.96} & 7.57
 & \bad{219} & \bad{151} \\

Lorenz96-Base
 & \first{5.19} & \first{1.33}
 & 8.03 & \second{2.97}
 & \second{6.35} & 2.49
 & 10.98 & 6.02
 & 17.42 & 3.41
 & 10.38 & 3.64
 & \bad{20.17} & \bad{15.56} \\

Lorenz96-Switch
 & \first{9.56} & \first{3.14}
 & \second{11.96} & 5.34
 & 10.73 & \second{4.73}
 & 13.68 & 7.80
 & 21.61 & 4.59
 & 11.78 & 5.25
 & \bad{24.42} & \bad{17.90} \\

Chua-Base
 & \second{0.056} & \second{0.033}
 & 0.094 & 0.046
 & 0.186 & 0.119
 & 0.720 & 0.507
 & 1.11 & 0.482
 & \first{0.051} & \first{0.029}
 & \bad{1.77} & \bad{1.67} \\

Chua-Param
 & \second{0.021} & \second{0.011}
 & \first{0.013} & \first{0.008}
 & 0.097 & 0.078
 & 0.681 & 0.559
 & 0.944 & 0.353
 & 0.030 & 0.021
 & \bad{2.31} & \bad{2.14} \\

Chua-Switch
 & 0.178 & 0.106
 & \second{0.174} & \second{0.123}
 & 0.318 & 0.230
 & 0.770 & 0.591
 & 1.11 & 0.584
 & \first{0.099} & \first{0.068}
 & \bad{1.74} & \bad{1.64} \\

\addlinespace[2pt]
\multicolumn{15}{l}{\emph{Switching Linear Dynamical System}} \\
SLDS-Base
 & 2.84 & 1.46
 & 4.54 & 2.80
 & \second{2.27} & \first{1.05}
 & 4.42 & 3.50
 & 2.96 & 1.96
 & \first{1.96} & \second{1.10}
 & 3.52 & 2.97 \\

SLDS-Param
 & 2.36 & \second{1.36}
 & 2.60 & 1.58
 & \first{2.18} & \first{0.91}
 & 2.26 & 1.50
 & \second{2.19} & 1.38
 & 2.30 & 1.80
 & 2.57 & 2.13 \\

SLDS-Switch
 & \first{4.05} & \first{2.02}
 & 7.84 & 4.84
 & 9.56 & 5.19
 & 4.73 & \second{3.38}
 & \second{4.56} & 3.38
 & 9.47 & 5.82
 & 8.23 & 6.81 \\

\addlinespace[2pt]
\multicolumn{15}{l}{\emph{Seasonal AR Shocks (SAR)}} \\
SAR-Base
 & \first{0.055} & 0.011
 & \first{0.055} & \second{0.013}
 & 0.074 & 0.021
 & \second{0.056} & \first{0.010}
 & 0.056 & 0.013
 & \first{0.055} & 0.013
 & \bad{1.85} & \bad{1.21} \\

SAR-Param
 & \first{0.355} & \first{0.053}
 & \second{0.361} & \second{0.065}
 & 0.480 & 0.128
 & 0.380 & \first{0.053}
 & 0.366 & 0.075
 & 0.359 & 0.065
 & \bad{10.83} & \bad{7.43} \\

\addlinespace[2pt]
\multicolumn{15}{l}{\emph{GARCH Volatility Shocks}} \\
GARCH-Base
 & \first{0.227} & 0.220
 & 0.234 & \second{0.201}
 & 0.271 & \first{0.174}
 & 0.264 & 0.190
 & \second{0.255} & 0.217
 & 0.261 & 0.210
 & 0.255 & 0.208 \\

GARCH-Param
 & \first{0.177} & 0.174
 & \second{0.186} & \second{0.156}
 & 0.220 & 0.138
 & 0.183 & \first{0.135}
 & 0.191 & 0.172
 & 0.206 & 0.163
 & 0.236 & 0.196 \\

\addlinespace[2pt]
\multicolumn{15}{l}{\emph{Double-Well Potential (DW)}} \\
DW-Base
 & 0.054 & \first{0.028}
 & 0.059 & 0.043
 & 0.091 & 0.057
 & 0.055 & \second{0.034}
 & \first{0.049} & 0.042
 & \first{0.049} & 0.043
 & \bad{1.47} & \bad{1.41} \\

DW-Param
 & \first{0.682} & \first{0.506}
 & 1.08 & 0.843
 & 0.983 & \second{0.721}
 & \second{0.847} & 0.711
 & 1.03 & 0.856
 & 1.00 & 0.815
 & 0.837 & 0.756 \\

\addlinespace[2pt]
\multicolumn{15}{l}{\emph{Ornstein--Uhlenbeck (OU) Diffusions}} \\
OU-Base
 & \first{0.234} & 0.195
 & 0.251 & \second{0.131}
 & 0.273 & \first{0.112}
 & \first{0.234} & 0.166
 & 0.251 & 0.156
 & \second{0.241} & 0.178
 & 0.237 & 0.216 \\

OU-Param
 & \first{0.239} & 0.170
 & 0.251 & \second{0.131}
 & 0.273 & \first{0.112}
 & \first{0.239} & 0.163
 & 0.251 & 0.156
 & \second{0.241} & 0.178
 & 0.383 & 0.358 \\

\addlinespace[2pt]
\multicolumn{15}{l}{\emph{Real-World Benchmarks}} \\
ETTm1
 & \second{8.97} & \first{5.37}
 & 9.12 & 5.63
 & \first{8.83} & \second{5.49}
 & 9.71 & 6.25
 & 9.22 & 5.69
 & 11.14 & 7.11
 & \bad{43.69} & \bad{34.25} \\

ETTh1
 & 10.97 & 5.75
 & \second{10.51} & 5.23
 & 10.97 & \first{4.83}
 & \first{10.39} & \second{5.00}
 & 10.75 & 5.55
 & 14.52 & 9.01
 & \bad{---} & \bad{---} \\


\bottomrule
\end{tabular}
\caption{\textbf{Stress-testing with stochastic and chaotic systems and controlled non-stationarity.} Models: fr (\Fern), tm (TimeMixer), tst (PatchTST), dl (DLinear), kp (Koopa), mtcn (ModernTCN), pfnn (PFNN). Lower is better. \first{Purple}: best; \second{Light Purple}: second-best; \bad{Light Orange}: diverged; `---': catastrophic.}
\label{tab:synthetic}
\end{table*}
 
\paragraph{Rethinking Benchmarking}
Parallel to the \textit{Information Monotonicity Assumption} we have \textbf{Provenance-over-Structure Fallacy:} (1) Real-world non-stationarity is deemed too complex to simulate, yet structural generators are dismissed as 'toy', without specifying what dynamical properties the `real' data actually provides that the generator lacks. (2) By extension, real-world sourced datasets are treated as inherently superior---\textit{as if provenance were a scientific property}. (3) Benchmark breadth masquerades as scientific validity---but structurally similar datasets tested eight times is n=1 with extra steps. (4) Winning on \textit{one realization} of non-stationarity is mistaken for mastering non-stationarity itself.    

We take the opposite view to all four points: (1) Current benchmarks \textit{are} easy: extremely simple models rarely fail catastrophically. (2) Historical data is only one realized path among counterfactuals; we risk leaderboard-ism over \textbf{historical path emulation}, when we don't know if a 5\% better MSE is statistically meaningful; (3) Benchmarks \textit{behave} like quasi-periodic datasets; many SOTA models bake in strong seasonal/trend priors and frequency decompositions. We risk conflating \textit{one} process with `real-world complexity'. (4) Better MSE on artifact-riddled quasi-periodic data can't be credibly extrapolated to unfalsifiable nonstationarity claims. In summary, \textit{we should not mistake where data comes from for what it can tell you}.

Just like nonstationarity handling, the core task of long-horizon forecasting, \textbf{conditional manifold transport}, \textit{remains a merely plausible narrative} with current benchmarks until we \textit{specify a structural generator}. We should do that. \textbf{Doing so is strictly better:} DGP and shock semantics are known by construction, and claims about nonstationarity handling become \textit{falsifiable}. Nonstationarity is rarely mystical for industry practitioners, and most can \textit{articulate} what makes their domain's nonstationarity special---interaction structure, shock timing, regime persistence, seasonal coupling---with ease. By \textit{explicitly modelling} the nonstationary scenarios, we eliminate two types of model failures: (1) true negatives that can't even do well on clean and noise-free generated data and (2) false positives that game the `real' data but stumble immediately under stress.   
 
\begin{figure*}[t]
\centering
\includegraphics[width=\textwidth]{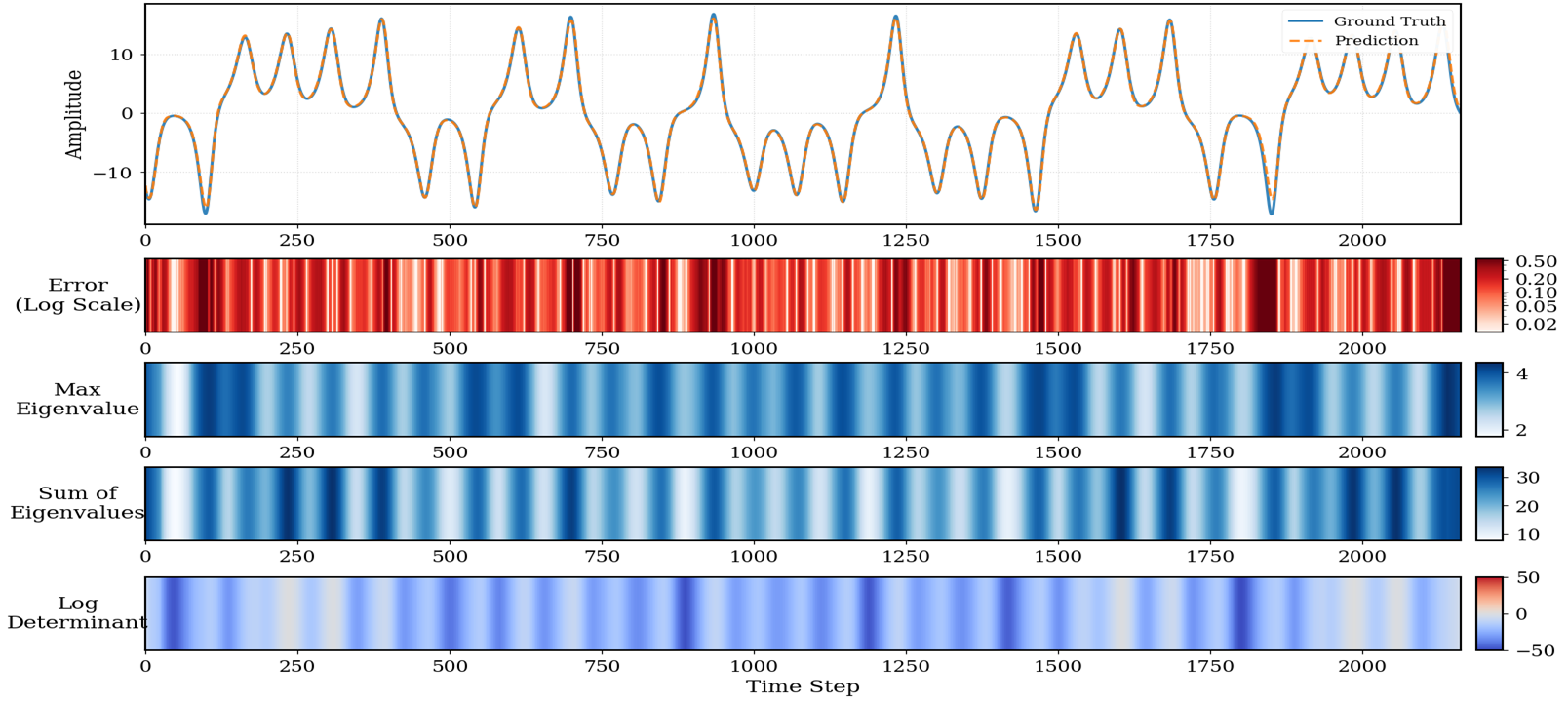} 
\caption{Diagnostics for Lorenz-63 x. Top to bottom: forecast (orange) vs. ground truth (blue); log absolute error; mean patch-wise max eigenvalue; mean patch-wise sum of eigenvalues; mean patch-wise log of products of eigenvalues.}
\label{fig:diag_plot}
\end{figure*}

\paragraph{The Gold-Standard: Synthetic Benchmarking}
If, with Arnold, \textit{mathematics is the part of physics where experiments are cheap}, then \textbf{\textit{synthetic data is forecasting where the ground truth is actually true}}. Dismissing these controlled environments as `not real' is akin to dismissing wind tunnel air as `not real wind'. By picking systems with attractors and structured invariant measures, \textbf{the problems with real-world datasets vanish}: (1) no measurement errors and data-quality artifacts; (2) distributional properties are analytical, not assumed or inferred; (3) random walk and non-dissipative systems are ruled out, risks of measuring another \texttt{Exchange} avoided; (4) non-stationary shock is defined by the user, recency bias between validation and train is eliminated unless we deliberately allow it.  

Non-stationarity handling becomes a \textbf{falsifiable claim} with experimental intervention where the \textbf{type, timing, magnitude, and before/after distributions} of shocks are all explicit and precise. We design three controlled non-stationary scenarios, \textbf{all occurring exactly at the midpoint of the \textit{training} trajectory}: (1) \textbf{Parameter drift:} system parameters shift slightly (same initial conditions); (2) \textbf{State perturbation:} state variables receive an additive shock (same parameters); (3) \textbf{Regime replacement:} the trajectory switches to a different one (different parameters \textit{and} initial conditions). Exact numerical parameters are in Appendix~\ref{tab:app:shock-params}). In our view, \textbf{synthetic benchmarks are software tests for forecasters}---refusing them is like skipping sensible edge-case tests in favour of a decades-old test suite from a different domain and calling it your `real-world' CI/CD.

Table~\ref{tab:synthetic} (shock table) evaluates \Fern against five LTSF baselines and one invariant learner: DLinear~\cite{zeng2023transformersAAAI}, TimeMixer\cite{wang2024timemixer}, PatchTST\cite{nie2023patchtst}, Koopa~\cite{liu2023koopa}, ModernTCN~\cite{luo2024moderntcn} and PFNN~\cite{cheng2025linearchaos} on 21 synthetic dynamical systems and 2 ETTs using 336-step input and 336-step forecast horizons, averaging results over 2 random seeds. Tables~\ref{tab:main_results} present tests with 4 random seed, 4 prediction horizons (96, 192, 336, 720) averaged results on \Fern and TimeMixer, PatchTST and DLinear with standard errors on MSE. We apply a 3-epoch grace period in Table~\ref{tab:synthetic} but not Table~\ref{tab:main_results} for a minimal of recency bias w.r.t ETT. Setting in Appendix~\ref{app:exp_setup}, Appendix~\ref{app:lit:recency}.

Table.~\ref{tab:synthetic} shows that among our choices, the \textit{diffusively blurred geometry} of a stochastic system poses far fewer threats compared to non-stationary shocks or chaotic systems, whose challenge is to \textit{locate relevant portion of a clear chaotic manifold}. Simple models such as DLinear and Koopa ranked 1st or 2nd on some stochastic systems and the (less problematic) ETT1. Yet, on Rossler, one of the easier chaotic systems, they show MSE of 5.42 and 11.94 vs \Fern's 0.019, errors 285× and 628× larger, and it worsens to 790× and 714× under parameter shift. The astonishing degree of brittleness negates any illusion for them to serve as general forecasters, especially given the prevalence of short term chaos in real life. After all, forecasting is an aid to \textbf{decision making}: worst-case robustness is important, if not more important than mean matching.
   
In stark contrast, \Fern delivers a sweeping victory: best MSE/WD in 19 of 21 scenarios, with a staggering 98\% lower MSE than TimeMixer on Rössler. On Lorenz-63, baselines collapse to mean-guessing early---DLinear at horizon 96, TimeMixer/PatchTST at 192 (Table~\ref{app:table:lorenz})---while \Fern maintains pointwise accuracy until horizon 720. At 720 steps---roughly 6.5 Lyapunov times, where errors amplify $\approx 650\times$ and pointwise prediction is provably impossible---\textbf{geometry persists}: SWD 4.89 versus 10--40 for baselines.

Fig.~\ref{fig:diag_plot} shows the \textbf{eigen-profile} of \Fern. When MSE is low ($1/8$ of SD), they still reveal additional insights: (1) \textbf{spectral radius} spikes at violent lobe-switching flips, coinciding with the largest errors---the model `believes' these segments are hardest and gets validation; (2) \textbf{trace} spikes within-lobe, suggesting less dominant eigen-directions activate during stable traversal; (3) \textbf{log determinant} captures volume scaling. Notably, some lowest-error regions (pale, row 2) coincide with low trace and strongly negative log-determinant \textit{immediately before} error spikes. This suggests \textbf{a specific failure mode}: the model becomes overconfident in its mean prediction, collapses variance, and consequently suffers when the system transitions unexpectedly.

\begin{table}[t]
\centering
\small
\setlength{\tabcolsep}{4pt}
\begin{tabular}{lrr}
\toprule
\textbf{Variant} & \textbf{H1 MSE↓} & \textbf{L63 MSE↓} \\
\midrule
Base              & 10.96            & \textbf{21.66} \\
No enc.\,+ no $\mu$ upd. 
                  & --               & 194.43 \\
Only encoder      & 11.17            & 27.09 \\
No rotation       & 11.84            & 27.62 \\
No patch          & 10.99            & 22.86 \\
Reflections = 2   & 11.52            & 26.14 \\
Reflections = 24 (8-block) 
                  & \textbf{10.91}   & 23.92 \\
\bottomrule
\end{tabular}
\caption{\textbf{Ablations (MSE only).} H1 = ETTh1, L63 = Lorenz-63, pred.\ length 192. 
Full metrics (MAE etc) reported in Appendix~\ref{tab:app:ablations-192-stacked}.}
\label{tab:ablations-192-main}
\end{table}

We provide simple \textbf{ablation summary} in Table.~\ref{tab:ablations-192-main} and in Appendix full details in Table~\ref{tab:app:ablations-192-stacked}. We find:
(i) Bidirectional encoder successfully compresses the input information into the latent space; without encoder, the model is useless, MSE and SWD explode. 
(ii) In the `only encoder' case, we disable the repeated translation in the decoder. This hurts ETTh1 and, more strongly, Lorenz, suggesting that the data-dependent translations are particularly important for chaotic systems;
(iii) removing either the eigen-based rotation or the patching mechanism consistently degrades Lorenz performance and often harms ETTh1, confirming that local geometric alignment and patch-wise conditioning are both doing real work; and
(iv) increasing the number of reflections from 0 (no rotation) to 24 \textbf{monotonically} improves MSE, SWD, and EPT on all three datasets at a modest runtime cost, making the 24-reflection configuration the strongest variant in this ablation grid. This suggests that a full-complexity spectral map \textit{is} important, and indirectly confirms the importance of patching, which makes a low-cost search possible.

\section{Literature Review and Discussion}
The LTSF field has been dominated by Transformer variants \citep{zhang2023crossformer,liu2024itransformer}. Foundational critiques \citep{zeng2023transformersAAAI,bergmeir2023foresight} on model complexity and evaluation paradigms catalyzed a shift to simplicity and interpretability. Recent efforts emphasize linearity and efficiency \citep{xu2024fits,yang2024fasttf, huang2025timebase}, and frequency-domain analysis for periodic signals \citep{wang2024fredf,wu2023timesnet}. \textbf{Koopman operator theory}~\citep{brunton2022modernkoopman}, is another direction with physics-informed design priors~\citep{liu2023koopa,zhang2025skolr}. Emerging focus on endogenous/exogenous structure~\citep{qiu2025dag,wang2024timexer} also provides a complementary angle to our discussion. Our ellipsoidal method can be extended to conformal prediction like \cite{xu2024multidimspci} in future work. \textit{Crucially}, our \textit{conditional} task \textit{differs from learning \textbf{global} invariants} (e.g., Koopman operators or Neural ODEs); we clarify scope in App.~\ref{app:lit:scope} and show one global learner's catastrophic divergence on conditional tasks in main text, with discussion in App.~\ref{app:lit:MNO}.

We propose measuring WD and EPT with technical details in App.~\ref{app:lit:metrics}. Pointwise metrics penalize phase shifts: $[0, 1, 2, 3]$ and $[1, 2, 3, 4]$ may be intuitively similar, as are $[0, 2, 4, 6]$ and $[2, 4, 6, 8]$, yet the same one-step shift yields $4 \times$ MSE. They also leave \textit{no room for error}: correct predictions that arrive too early or too late are treated as entirely wrong. DTW~\citep{sakoe1978dtw} partially addresses phase but is computationally heavy and non-differentiable. The \textbf{Wasserstein-2 (W2)} distance provides an alternative: it measures the minimal ``work'' to morph one distribution into another, rewarding sharp forecasts even under phase shift. Our 1D W2 proposal coincides with ideas independently developed in several fields
\citep{MuskulusVerduynLunel2011, Wiesel2022, AounEtAl2024,
BotvinickGreenhouseOpreaMaulikYang2024, wang2025distdf}, suggesting a convergent consensus. 

\noindent\textbf{Code availability.} An anonymized repository containing the implementation and supplemental materials is available at \url{https://anonymous.4open.science/r/FernPaper-58B4}.

\clearpage
\section*{Impact Statement}
This paper presents work whose goal is to advance the field of Machine
Learning. There are many potential societal consequences of our work, none
which we feel must be specifically highlighted here.

\clearpage
\bibliography{example_paper}

@article{brenier1991polar,
  title        = {Polar factorization and monotone rearrangement of vector-valued functions},
  author       = {Brenier, Yann},
  journal      = {Communications on Pure and Applied Mathematics},
  volume       = {44},
  number       = {4},
  pages        = {375--417},
  year         = {1991},
  publisher    = {Wiley}
}

@article{peyre2019computational,
  title        = {Computational Optimal Transport},
  author       = {Peyr{\'e}, Gabriel and Cuturi, Marco},
  journal      = {Foundations and Trends in Machine Learning},
  volume       = {11},
  number       = {5--6},
  pages        = {355--607},
  year         = {2019},
  publisher    = {Now Publishers}
}

@article{bonneel2015sliced,
  title     = {Sliced \& Radon Wasserstein Barycenters of Measures},
  author    = {Bonneel, Nicolas and Rabin, Julien and Peyr{\'e}, Gabriel and Pfister, Hanspeter},
  journal   = {Journal of Mathematical Imaging and Vision},
  volume    = {51},
  number    = {1},
  pages     = {22--45},
  year      = {2015},
  doi       = {10.1007/s10851-014-0506-3},
  publisher = {Springer}
}

@article{sakoe1978dtw,
  title        = {Dynamic programming algorithm optimization for spoken word recognition},
  author       = {Sakoe, Hiroaki and Chiba, Seibi},
  journal      = {IEEE Transactions on Acoustics, Speech, and Signal Processing},
  volume       = {26},
  number       = {1},
  pages        = {43--49},
  year         = {1978}
}

@article{lorenz1963deterministic,
  title        = {Deterministic Nonperiodic Flow},
  author       = {Lorenz, Edward N.},
  journal      = {Journal of the Atmospheric Sciences},
  volume       = {20},
  number       = {2},
  pages        = {130--141},
  year         = {1963},
  doi          = {10.1175/1520-0469(1963)020<0130:DNF>2.0.CO;2}
}

@article{brunton2017havok,
  author  = {Brunton, Steven L. and Brunton, Bingni W. and Proctor, Joshua L. and Kaiser, Eurika and Kutz, J. Nathan},
  title   = {Chaos as an Intermittently Forced Linear System},
  journal = {Nature Communications},
  volume  = {8},
  number  = {1},
  pages   = {19},
  year    = {2017},
  doi     = {10.1038/s41467-017-00030-8},
}

@article{huang2020anf,
  title        = {Augmented Normalizing Flows: Bridging the Gap Between Generative Flows and Latent Variable Models},
  author       = {Huang, Chin-Wei and Dinh, Laurent and Courville, Aaron},
  journal      = {arXiv preprint arXiv:2002.07101},
  year         = {2020}
}

@inproceedings{ko2023homotopy,
  title        = {Homotopy-based Training of NeuralODEs for Accurate Dynamics Discovery},
  author       = {Ko, Joon-Hyuk and Koh, Hankyul and Park, Nojun and Jhe, Wonho},
  booktitle    = {Advances in Neural Information Processing Systems (NeurIPS)},
  year         = {2023}
}

@inproceedings{jiang2023neuraloperators,
  title        = {Training Neural Operators to Preserve Invariant Measures of Chaotic Attractors},
  author       = {Jiang, Ruoxi and Lu, Peter Y. and Orlova, Elena and Willett, Rebecca},
  booktitle    = {Advances in Neural Information Processing Systems (NeurIPS)},
  year         = {2023}
}

@article{schiff2024dyslim,
  title        = {DySLIM: Dynamics-Stable Learning by Invariant Measure for Chaotic Systems},
  author       = {Schiff, Yair and Wan, Zhong Yi and Parker, Jeffrey B. and Hoyer, Stephan and Kuleshov, Volodymyr and Sha, Fei and Zepeda-N{\'u}{\~n}ez, Leonardo},
  journal      = {arXiv preprint arXiv:2402.04467},
  year         = {2024}
}

@article{bergmeir2023foresight,
  author  = {Bergmeir, Christoph},
  title   = {Common Pitfalls and Better Practices in Forecast Evaluation for Data Scientists},
  journal = {Foresight: The International Journal of Applied Forecasting},
  number  = {70},
  pages   = {5--12},
  year    = {2023},
  url     = {https://cbergmeir.com/publications/2023-01-01_bergmeir2023common/}
}

@article{brunton2016sindy,
  author  = {Brunton, Steven L. and Proctor, Joshua L. and Kutz, J. Nathan},
  title   = {Discovering Governing Equations from Data by Sparse Identification of Nonlinear Dynamical Systems},
  journal = {Proceedings of the National Academy of Sciences},
  volume  = {113},
  number  = {15},
  pages   = {3932--3937},
  year    = {2016},
  doi     = {10.1073/pnas.1517384113}
}

@article{brunton2022modernkoopman,
  author  = {Brunton, Steven L. and Budi{\v{s}}i{\'c}, Marko and Kaiser, Eurika and Kutz, J. Nathan},
  title   = {Modern Koopman Theory for Dynamical Systems},
  journal = {SIAM Review},
  volume  = {64},
  number  = {2},
  pages   = {229--340},
  year    = {2022},
  doi     = {10.1137/21M1401243}
}

@article{conti2024venivindyvici,
  author  = {Conti, Paolo and Kneifl, Jonas and Manzoni, Andrea and Frangi, Attilio and Fehr, J{\"o}rg and Brunton, Steven L. and Kutz, J. Nathan},
  title   = {VENI, VINDy, VICI: A Variational Reduced-Order Modeling Framework with Uncertainty Quantification},
  journal = {arXiv},
  year    = {2024},
  eprint  = {2405.20905},
  url     = {https://arxiv.org/abs/2405.20905}
}

@inproceedings{nie2023patchtst,
  author    = {Nie, Yao and Nguyen, Nghia Hoang and Sinthong, Phanwadee and Kalagnanam, Jayant},
  title     = {A Time Series is Worth 64 Words: Long-term Forecasting with Transformers},
  booktitle = {International Conference on Learning Representations (ICLR)},
  year      = {2023},
  url       = {https://openreview.net/forum?id=Jbdc0vTOcol}
}

@article{Householder1958,
  author  = {Householder, Alston S.},
  title   = {Unitary Triangularization of a Nonsymmetric Matrix},
  journal = {Journal of the ACM},
  volume  = {5},
  number  = {4},
  pages   = {339--342},
  year    = {1958}
}

@book{Golub1996Matrix,
  author    = {Golub, Gene H. and Van Loan, Charles F.},
  title     = {Matrix Computations},
  edition   = {3rd},
  publisher = {Johns Hopkins University Press},
  address   = {Baltimore, MD},
  year      = {1996}
}

@incollection{takens1981detecting,
  author    = {Takens, Floris},
  title     = {Detecting Strange Attractors in Turbulence},
  booktitle = {Dynamical Systems and Turbulence, Warwick 1980},
  series    = {Lecture Notes in Mathematics},
  volume    = {898},
  pages     = {366--381},
  publisher = {Springer},
  year      = {1981},
  doi       = {10.1007/BFb0091924}
}

@article{Klower2023LowPrecChaos,
  title   = {Periodic orbits in chaotic systems simulated at low precision},
  author  = {Kl{\"o}wer, Milan and Coveney, Peter V. and Paxton, E. Adam and Palmer, Tim N.},
  journal = {Scientific Reports},
  volume  = {13},
  number  = {11410},
  year    = {2023},
  doi     = {10.1038/s41598-023-37004-4}
}

@article{Teixeira2007TimeStep,
  title   = {Time Step Sensitivity of Nonlinear Atmospheric Models: Numerical Convergence, Truncation Error Growth, and Ensemble Design},
  author  = {Teixeira, Jo{\~a}o and Reynolds, Carolyn A. and Judd, Kevin},
  journal = {Journal of the Atmospheric Sciences},
  volume  = {64},
  number  = {1},
  pages   = {175--189},
  year    = {2007},
  doi     = {10.1175/JAS3824.1}
}

@article{MuskulusVerduynLunel2011,
  title   = {Wasserstein distances in the analysis of time series and dynamical systems},
  author  = {Muskulus, Michael and Verduyn Lunel, Sjoerd},
  journal = {Physica D: Nonlinear Phenomena},
  volume  = {240},
  number  = {1},
  pages   = {45--58},
  year    = {2011},
  issn    = {0167-2789},
  doi     = {10.1016/j.physd.2010.08.005}
}

@article{Wiesel2022,
  title   = {Measuring association with Wasserstein distances},
  author  = {Wiesel, Johannes C. W.},
  journal = {Bernoulli},
  volume  = {28},
  number  = {4},
  pages   = {2816--2832},
  year    = {2022},
  doi     = {10.3150/21-BEJ1438},
  url     = {https://projecteuclid.org/journals/bernoulli/volume-28/issue-4/Measuring-association-with-Wasserstein-distances/10.3150/21-BEJ1438.full}
}

@article{AounEtAl2024,
  title   = {Beyond correlation: optimal transport metrics for characterizing representational stability and remapping in neurons encoding spatial memory},
  author  = {Aoun, Andrew and Shetler, Oliver and Raghuraman, Radha and Rodriguez, Gustavo A. and Hussaini, S. Abid},
  journal = {Frontiers in Cellular Neuroscience},
  volume  = {17},
  pages   = {1273283},
  year    = {2024},
  note    = {eCollection 2023},
  doi     = {10.3389/fncel.2023.1273283}
}

@misc{BotvinickGreenhouseOpreaMaulikYang2024,
  title         = {Measure-Theoretic Time-Delay Embedding},
  author        = {Botvinick-Greenhouse, Jonah and Oprea, Maria and Maulik, Romit and Yang, Yunan},
  year          = {2024},
  eprint        = {2409.08768},
  archivePrefix = {arXiv},
  primaryClass  = {math.DS},
  url           = {https://arxiv.org/abs/2409.08768}
}

@article{Liao2014CNS,
  title   = {On the mathematically reliable long-term simulation of chaotic solutions of Lorenz equation in the interval [0, 10000]},
  author  = {Liao, Shijun and Wang, Pengfei},
  journal = {Science China Physics, Mechanics \& Astronomy},
  volume  = {57},
  number  = {2},
  pages   = {330--335},
  year    = {2014},
  doi     = {10.1007/s11433-013-5374-2}
}

@article{wangli2024dsdl,
  author  = {Wang, Min and Li, Jin},
  title   = {Interpretable Predictions of Chaotic Dynamical Systems Using Dynamical System Deep Learning},
  journal = {Scientific Reports},
  volume  = {14},
  number  = {1},
  pages   = {3143},
  year    = {2024},
  doi     = {10.1038/s41598-024-27171-9}
}

@article{Stark1999,
  author    = {J. Stark and D. S. Broomhead and M. E. Davies and J. Huke},
  title     = {Delay embeddings for forced systems. I. Deterministic forcing},
  journal   = {Journal of Nonlinear Science},
  volume    = {9},
  number    = {3},
  pages     = {255--332},
  year      = {1999},
  doi       = {10.1007/s003329900078}
}

@article{Stark2003,
  author    = {J. Stark},
  title     = {Delay embeddings for forced systems. II. Stochastic forcing},
  journal   = {Journal of Nonlinear Science},
  volume    = {13},
  number    = {5},
  pages     = {519--577},
  year      = {2003},
  doi       = {10.1007/s00332-003-0534-4}
}

@article{wang2024timemixer,
  author  = {Wang, Shiyu and Wu, Haixu and Shi, Xiaoming and Hu, Tengge and Luo, Huakun and Ma, Lintao and Zhang, James Y. and Zhou, Jun},
  title   = {TimeMixer: Decomposable Multiscale Mixing for Time Series Forecasting},
  journal = {arXiv},
  year    = {2024},
  eprint  = {2405.14616},
  url     = {https://arxiv.org/abs/2405.14616}
}

@article{zeng2023transformersAAAI,
  author  = {Zeng, Ailing and Chen, Muxi and Zhang, Lei and Xu, Qiang},
  title   = {Are Transformers Effective for Time Series Forecasting?},
  journal = {Proceedings of the AAAI Conference on Artificial Intelligence},
  volume  = {37},
  number  = {9},
  pages   = {11121--11128},
  year    = {2023},
  doi     = {10.1609/aaai.v37i9.26323}
}

@inproceedings{liu2023koopa,
  title     = {Koopa: Learning Non-stationary Time Series Dynamics with Koopman Predictors},
  author    = {Liu, Yong and Li, Chenyu and Wang, Jianmin and Long, Mingsheng},
  booktitle = {Advances in Neural Information Processing Systems},
  year      = {2023},
  url       = {https://arxiv.org/abs/2305.18803}
}

@inproceedings{zhang2025skolr,
  title     = {SKOLR: Structured Koopman Operator Linear RNN for Time‑Series Forecasting},
  author    = {Zhang, Yitian and Ma, Liheng and Valkanas, Antonios and Oreshkin, Boris N. and Coates, Mark},
  booktitle = {Proceedings of the 42nd International Conference on Machine Learning (ICML)},
  year      = {2025},
  url       = {https://openreview.net/forum?id=Xg1BGlybfq}
}

@inproceedings{huang2025timebase,
  title     = {TimeBase: The Power of Minimalism in Efficient Long-term Time Series Forecasting},
  author    = {Huang, Qihe and Zhou, Zhengyang and Yang, Kuo and Yi, Zhongchao and Wang, Xu and Jiang, Wu and Wang, Yang},
  booktitle = {Proceedings of the 42nd International Conference on Machine Learning (ICML)},
  year      = {2025},
  url       = {https://proceedings.mlr.press/v267/huang25a.html}
}

@misc{yang2024fasttf,
  title  = {FastTF: 4 Parameters are All You Need for Long-term Time Series Forecasting},
  author = {Yang, Chuhong and Qi, Yuanjing and Li, Bin and Wu, Nan},
  year   = {2024},
  note   = {Submitted to ICLR 2025 (OpenReview)},
  url    = {https://openreview.net/forum?id=CZiP7GpmX7}
}

@inproceedings{xu2024fits,
  author    = {Xu, Zhijian and Zeng, Ailing and Xu, Qiang},
  title     = {FITS: Modeling Time Series with 10\,K Parameters},
  booktitle = {Proceedings of the 12\textsuperscript{th} International Conference on Learning Representations (ICLR)},
  year      = {2024},
  url       = {https://openreview.net/forum?id=bWcnvZ3qMb}
}

@inproceedings{wu2023timesnet,
  author    = {Wu, Haixu and Hu, Tengge and Liu, Yong and Zhou, Hang and Wang, Jianmin and Long, Mingsheng},
  title     = {TimesNet: Temporal 2D-Variation Modeling for General Time Series Analysis},
  booktitle = {Proceedings of the 11th International Conference on Learning Representations (ICLR)},
  year      = {2023},
  url       = {https://openreview.net/forum?id=ju_Uqw384Oq}
}

@inproceedings{liu2024itransformer,
  author    = {Liu, Yong and Hu, Tengge and Zhang, Haoran and Wu, Haixu and Wang, Shiyu and Ma, Lintao and Long, Mingsheng},
  title     = {iTransformer: Inverted Transformers Are Effective for Time Series Forecasting},
  booktitle = {Proceedings of the 12\textsuperscript{th} International Conference on Learning Representations (ICLR)},
  year      = {2024},
  url       = {https://openreview.net/forum?id=JePfAI8fah}
}

@article{sutton2019bitterlesson,
  author  = {Richard S. Sutton},
  title   = {The Bitter Lesson},
  journal = {Incomplete Ideas (blog)},
  year    = {2019},
  note    = {Essay, accessed 2025},
  url     = {http://incompleteideas.net/IncIdeas/BitterLesson.html}
}

@inproceedings{zhang2023crossformer,

author = {Zhang, Yunhao and Yan, Junchi},

title = {Crossformer: Transformer Utilizing Cross-Dimension Dependency for Multivariate Time Series Forecasting},

booktitle = {Proceedings of the 11th International Conference on Learning Representations (ICLR)},

year = {2023},

url = {https://openreview.net/forum?id=vSVLM2j9eie}

}

@inproceedings{li2022learning,
  title     = {Learning Dissipative Dynamics in Chaotic Systems},
  author    = {Li, Zongyi and Liu-Schiaffini, Miguel and Kovachki, Nikola and Liu, Burigede and Azizzadenesheli, Kamyar and Bhattacharya, Kaushik and Stuart, Andrew and Anandkumar, Anima},
  booktitle = {Advances in Neural Information Processing Systems},
  year      = {2022}
}

@inproceedings{loshchilov2017decoupled,
  title     = {Decoupled Weight Decay Regularization},
  author    = {Loshchilov, Ilya and Hutter, Frank},
  booktitle = {International Conference on Learning Representations (ICLR)},
  year      = {2019},
  note      = {First formal description of AdamW optimization method}
}

@book{butcher1987numerical,
  title     = {The Numerical Analysis of Ordinary Differential Equations: Runge–Kutta and General Linear Methods},
  author    = {Butcher, John C.},
  publisher = {John Wiley \& Sons},
  year      = {1987},
  isbn      = {978-0471607856},
  note      = {Comprehensive treatment of RK methods and Butcher’s formalism}
}

@inproceedings{luo2024moderntcn,
  title     = {ModernTCN: A Modern Pure Convolution Structure for General Time Series Analysis},
  author    = {Luo, Donghao and Wang, Xue},
  booktitle = {The Twelfth International Conference on Learning Representations},
  year      = {2024},
  note      = {ICLR 2024 Spotlight},
  url       = {https://openreview.net/forum?id=vpJMJerXHU}
}

@article{cheng2025linearchaos,
  author    = {Xiaoyuan Cheng and Yi He and Yiming Yang and Xiao Xue
               and Sibo Cheng and Daniel Giles and Xiaohang Tang and Yukun Hu},
  title     = {Learning Chaos in a Linear Way},
  journal   = {arXiv preprint arXiv:2503.14702},
  year      = {2025}
}

@article{wang2024fredf,
  title         = {FreDF: Learning to Forecast in the Frequency Domain},
  author        = {Wang, Hao and Pan, Licheng and Chen, Zhichao and Yang, Degui and Zhang, Sen and Yang, Yifei and Liu, Xinggao and Li, Haoxuan and Tao, Dacheng},
  journal       = {arXiv preprint arXiv:2402.02399},
  year          = {2024},
  doi           = {10.48550/arXiv.2402.02399},
  url           = {https://arxiv.org/abs/2402.02399}
}

@article{qiu2025dag,
  title        = {{DAG}: A Dual Causal Network for Time Series Forecasting with Exogenous Variables},
  author       = {Qiu, Xiangfei and Zhu, Yuhan and Li, Zhengyu and Cheng, Hanyin and Wu, Xingjian and Guo, Chenjuan and Yang, Bin and Hu, Jilin},
  journal      = {arXiv preprint arXiv:2509.14933},
  year         = {2025},
  doi          = {10.48550/arXiv.2509.14933}
}

@article{ChorinHaldKupferman2002OptimalPredictionMemory,
  author  = {Chorin, Alexandre J. and Hald, Ole H. and Kupferman, Raz},
  title   = {Optimal prediction with memory},
  journal = {Physica D: Nonlinear Phenomena},
  year    = {2002},
  volume  = {166},
  number  = {3-4},
  pages   = {239--257},
  doi     = {10.1016/S0167-2789(02)00446-3}
}

@article{wang2025distdf,
  title         = {DistDF: Time-Series Forecasting Needs Joint-Distribution Wasserstein Alignment},
  author        = {Wang, Hao and Pan, Licheng and Lu, Yuan and Chu, Zhixuan and Li, Xiaoxi and He, Shuting and Chen, Zhichao and Li, Haoxuan and Wen, Qingsong and Lin, Zhouchen},
  journal       = {arXiv preprint arXiv:2510.24574},
  year          = {2025},
  month         = oct,
  eprint        = {2510.24574},
  archivePrefix = {arXiv},
  primaryClass  = {cs.LG},
  doi           = {10.48550/arXiv.2510.24574},
  url           = {https://arxiv.org/abs/2510.24574}
}

@article{wang2024timexer,
  title         = {TimeXer: Empowering Transformers for Time Series Forecasting with Exogenous Variables},
  author        = {Wang, Yuxuan and Wu, Haixu and Dong, Jiaxiang and Qin, Guo and Zhang, Haoran and Liu, Yong and Qiu, Yunzhong and Wang, Jianmin and Long, Mingsheng},
  journal       = {arXiv preprint arXiv:2402.19072},
  year          = {2024},
  month         = feb,
  eprint        = {2402.19072},
  archivePrefix = {arXiv},
  primaryClass  = {cs.LG},
  doi           = {10.48550/arXiv.2402.19072},
  url           = {https://arxiv.org/abs/2402.19072}
}

@article{xu2024multidimspci,
  title         = {Conformal prediction for multi-dimensional time series by ellipsoidal sets},
  author        = {Xu, Chen and Jiang, Hanyang and Xie, Yao},
  journal       = {arXiv preprint arXiv:2403.03850},
  year          = {2024},
  month         = mar,
  eprint        = {2403.03850},
  archivePrefix = {arXiv},
  primaryClass  = {stat.ML},
  doi           = {10.48550/arXiv.2403.03850},
  url           = {https://arxiv.org/abs/2403.03850}
}
\bibliographystyle{icml2026} 

\clearpage
\appendix
\section{Appendix}

\subsection{Distributional metrics}
\label{app:lit:metrics}

For 1D distributions, W2 reduces to the $L_2$ distance between sorted values~\cite{peyre2019computational}.  
In our LTSF setting, for a single channel and horizon length $H$ we treat
$y^\star = (y^\star_1,\dots,y^\star_H)$ and $y = (y_1,\dots,y_H)$ as $H$ scalar samples (ignoring time order) and compute
\[
W_2^2(\mu^\star,\mu) = \frac{1}{H}\sum_{h=1}^H \bigl(y^*_{(h)}-y_{(h)}\bigr)^2,
\]
where $y^*_{(1)}\le\cdots\le y^*_{(H)}$ and $y_{(1)}\le\cdots\le y_{(H)}$ are order statistics. This is \textbf{index-agnostic}: it compares histograms, ignoring temporal alignment. It complements MSE by asking: \textit{did the model predict the correct values, regardless of when?}

\textbf{Sliced Wasserstein Distance (SWD)} extends W2 to multivariate distributions by projecting horizon vectors in $\mathbb{R}^H$ onto $L$ random directions and averaging the resulting 1D W2 distances~\citep{bonneel2015sliced}. In our tables, \textbf{WD} denotes the 1D W2 metric (no projection), and \textbf{SWD} denotes sliced W2 with $L=500$ projections.

\paragraph{Effective prediction time (EPT).} We complement W2 with Effective Prediction Time (EPT), defined as the first forecast step at which the error exceeds one standard deviation of the training data. EPT quantifies reliability: if multiple models have EPT $\approx 190$ on a 336-step horizon, steps $t > 190$ should be treated more as failure-mode analysis than actionable signal. For chaotic systems, EPT measures the first horizon step where the absolute error exceeds a 1$\sigma$ envelope.
Let $\epsilon_d$ be the training-set standard deviation for dimension $d$.
For each batch $b$ and dimension $d$,
\[
\mathrm{EPT}_{b,d}=\min\{s\in\{1,\dots,H\}: |y^{\mathrm{pred}}_{b,d,s}-y^{\mathrm{true}}_{b,d,s}|>\epsilon_d\},
\]
with $\mathrm{EPT}_{b,d}=H$ if the threshold is never exceeded, and report $\mathrm{EPT}_{\mathrm{avg}}=\frac{1}{BD}\sum_{b,d}\mathrm{EPT}_{b,d}$.

\paragraph{Householder-based orthogonal parameterization.}
We parameterize an orthogonal matrix as a product of $R$ Householder reflections
$H_i = I - 2 v_i v_i^\top$ with $\|v_i\|_2=1$, i.e.,
$U = H_R \cdots H_1$ \citep{Householder1958,Golub1996Matrix}.
This yields $U^\top U = I$ by construction; when $R$ is even, $\det(U)=+1$.

\subsection{App: Scope and Discussions}

\subsubsection{\Fern's role,  clarified}
\label{app:lit:scope}

\paragraph{What \Fern is, and what \Fern is not} Due to frequent misunderstanding about \Fern's role as a forecast model, we clarify:  \Fern is  \textbf{not an iterated one-step chaos specialist.} Such models recursively feed their own output to simulate long trajectories, and chaos-tuned variants (e.g., DSDL) excel on Lorenz dataset \citep{wangli2024dsdl}. \Fern is instead a direct, multi-step forecaster for general-purpose tasks, stress-tested on chaotic datasets only as a supplement to standard LTSF benchmarks. Also, the metric EPT we advocate for, is a common measure of one-step \textit{model stability} in chaotic dynamics, alongside variants like VPT (valid prediction time). We use EPT to measure \textit{dataset predictability}. Direct comparison between iterated and multi-step models can be misleading, as the latter are not optimized for it.

\Fern is \textbf{not a governing-equation or invariant-measure learner}. It aims at discrete-time conditional forecasting with data-driven spectral factors rather than system identification, distinct from (i) sparse governing equation finding SINDy/HAVOK/VINDy \citep{brunton2016sindy,brunton2017havok,conti2024venivindyvici} (ii) Neural-ODE identification \citep{ko2023homotopy} (iii) invariant-measure learner \citep{jiang2023neuraloperators, schiff2024dyslim}. 

\subsubsection{Markov Neural Operators}
\label{app:lit:MNO}
Operator-learning approaches (PFNN, MNO~\cite{li2022learning}) are \textit{invariant measure} learners of chaotic regimes, which pursue a complementary agenda: learning global evolution operators for specific systems. In contrast, \Fern targets local conditional geometry for generic forecasting tasks, prioritizing per-window spectral diagnostics over explicit operator recovery. 

However, many requests insist on probing \Fern's pointwise performances vs this class, and so we complied. In the main text, Table.~\ref{tab:synthetic} suggests PFNN performs reasonably on some smooth diffusion tasks (OU, DW), but often exhibits very large errors or outright divergence on chaotic and heavily forced settings (e.g., Rössler, Lorenz, ETTm1), under the unified training protocol. To understand the performance against data artifacts, we test on the ETTh2 case, and the reported error reaches $\mathcal{O}(10^9)$.

We were initially concerned about the errors and perform additional hyperparameter (learning rate, latent dimension size, network architecture etc) searches. Our implementation carefully follows the PFNN public repository. Based on \citep{cheng2025linearchaos}, the reported NRMSE is $0.49$ on $~\sim80000$ samples with $\mathrm{d}t=10^{-2}$ for Lorenz–63. Since the standard deviation of the Lorenz system is about $8$, this corresponds to roughly $15$–$20$ MSE over a $100$-step rollout. In contrast, FERN attains an MSE of about $1.2$ on $22$k samples with a $96$-step prediction horizon. We conclude that the reported error on $336$-step rollout is likely consistent with the paper.

\textbf{Why the difference with Markov System vs Conditional Forecasting?} Conceptually, LTSF models take a history window of shape $[\text{batch}, \text{feature}, \text{seq}_{0:100}]$ and predict a future window $[\text{batch}, \text{feature}, \text{seq}_{100:200}]$, often with \textit{channel-independent} parametrization, whereas Markov(-type) operators act along the temporal axis, mapping $[\text{batch}, \text{seq}_{0:1}, \text{feature}]$ to $[\text{batch}, \text{seq}_{1:2}, \text{feature}]$ by evolving the state one step at a time using \textit{only} the \textit{current state} information.

Lorenz etc.\textit{are} Markov systems but empirically \textbf{past information helps localize the system on its attractor}. The task of \emph{MNO} is to \emph{learn the global invariant measure} based on \emph{current} state, which is a \textbf{strictly more difficult problem} than \emph{learning conditional forecast distribution} with recent information. Lorenz63 has 3 state variables, $\sim 10$ variables are enough to invoke Takens' Embedding Theorem, so the remaining input length serves as historical context. We view this not as a flaw in PFNN per se, but as evidence that architectures tuned for Markovian operator learning do not transfer automatically to windowed multi-horizon forecasting; conversely, \Fern does not currently target PDE operators and should not be read as a replacement in that regime. 

\subsection{Additional Discussion} 
\label{app:lit:etth2}

\paragraph{Zero-inflation analysis.}
Tables~\ref{app:table:etth2-zero-patterns}, \ref{app:table:etth1-zero-patterns}, and \ref{app:table:weather-zero-patterns} detail zero patterns. ETTh2 and Weather are heavily affected; ETTh1 is better behaved. Traffic and Electricity show similar issues (analysis omitted for space).

\paragraph{Preprocessing policy.}
\label{app:lit:preprocessing}
ETTh2/m2 are unfixable via imputation. Our policy is a preliminary proposal and we leave refinement to future work:
\begin{itemize}
    \item Sentinel values $>10\%$ or zeros $>15\%$: drop column.
    \item Zeros $10$--$15\%$: apply $\mathrm{asinh}$ transform.
    \item Longest zero run $>1$ week: drop column.
    \item Longest zero run $>3$ hours: delete affected rows.
    \item Longest zero run $\le 3$ hours: forward--backward fill.
\end{itemize}
We apply this \textbf{\textit{only}} to ETTh1/m1 in the shock tables (see App.~\ref{app:exp_setup}). These datasets are well-behaved, so \textbf{\textit{no} columns is dropped and only zero imputation and row removals are performed.}

\begin{table}[ht]
\centering
\footnotesize 
\begin{tabular}{lrrrrl}
\toprule
Column & Total Zeros & \% Zeros & Isolated & Clustered  \\
\midrule
HUFL & 58   & 0.33\% & 1   & 57    \\
HULL & 3836 & 22.02\%& 163 & 3673  \\
MUFL & 0    & 0.00\% & 0   & 0     \\
MULL & 1067 & 6.13\% & 245 & 822   \\
LUFL & 1188 & 6.82\% & 184 & 1004  \\
LULL & 5792 & 33.25\%& 414 & 5378  \\
OT   & 1    & 0.01\% & 1   & 0     \\
\bottomrule
\end{tabular}
\caption{Zero–inflation patterns in \texttt{ETTh2}. ``Clustered'' counts zeros that belong to runs of length $>1$ (consecutive zeros). Type~1 columns show substantial intermittent zero bursts; Type~2 suggests occasional missing entries.}
\label{app:table:etth2-zero-patterns}
\end{table}

\begin{table}[ht]
\centering
\footnotesize

\begin{tabular}{lrrrr}
\toprule
Column & Total Zeros & \% Zeros & Isolated & Clustered \\
\midrule
HUFL & 89  & 0.51\% & 31  & 58  \\
HULL & 410 & 2.35\% & 256 & 154 \\
MUFL & 97  & 0.56\% & 19  & 78  \\
MULL & 236 & 1.35\% & 137 & 99  \\
LUFL & 60  & 0.34\% & 1   & 59  \\
LULL & 212 & 1.22\% & 7   & 205 \\
OT   & 111 & 0.64\% & 28  & 83  \\
\bottomrule
\end{tabular}
\caption{Zero--inflation patterns in \texttt{ETTh1}. ``Clustered'' counts zeros that belong to runs of length $>1$ (consecutive zeros).}
\label{app:table:etth1-zero-patterns}
\end{table}

\begin{table}[h]
\centering
\footnotesize
\setlength{\tabcolsep}{3pt}
\renewcommand{\arraystretch}{0.95}

\begin{tabular}{p{2.2cm}rrrr}
\toprule
Column & Total Zeros & \% Zeros & Isolated & Clustered \\
\midrule
p (mbar) & 0 & 0.00\% & 0 & 0 \\
T (degC) & 6 & 0.01\% & 4 & 2 \\
Tpot (K) & 0 & 0.00\% & 0 & 0 \\
Tdew (degC) & 39 & 0.07\% & 39 & 0 \\
rh (\%) & 0 & 0.00\% & 0 & 0 \\
VPmax (mbar) & 0 & 0.00\% & 0 & 0 \\
VPact (mbar) & 0 & 0.00\% & 0 & 0 \\
VPdef (mbar) & 3026 & 5.74\% & 45 & 2981 \\
sh (g/kg) & 0 & 0.00\% & 0 & 0 \\
H2OC (mmol/mol) & 0 & 0.00\% & 0 & 0 \\
rho (g/m**3) & 0 & 0.00\% & 0 & 0 \\
wv (m/s) & 0 & 0.00\% & 0 & 0 \\
max. wv (m/s) & 1 & 0.00\% & 1 & 0 \\
wd (deg) & 1 & 0.00\% & 1 & 0 \\
rain (mm) & 50785 & 96.37\% & 177 & 50608 \\
raining (s) & 49316 & 93.59\% & 115 & 49201 \\
SWDR (W/m$^2$) & 24749 & 46.97\% & 21 & 24728 \\
PAR ($\mu$mol/m$^2$/s) & 24614 & 46.71\% & 0 & 24614 \\
max. PAR ($\mu$mol/m$^2$/s) & 24209 & 45.94\% & 0 & 24209 \\
Tlog (degC) & 0 & 0.00\% & 0 & 0 \\
OT & 0 & 0.00\% & 0 & 0 \\
\bottomrule
\end{tabular}
\caption{Zero--inflation patterns in \texttt{Weather}. ``Clustered'' counts zeros that belong to runs of length $>1$ (consecutive zeros).}
\label{app:table:weather-zero-patterns}
\end{table}

\subsubsection{Checkpoint Strategy} 
\label{app:lit:recency}
\textbf{The "Anti-Predictive" Early Stop.} Table~\ref{tab:recency-etth1} (using an earlier \Fern variant) demonstrates that early validation error is often \emph{anti-predictive} of final test error, rendering it an \textbf{unreliable proxy} for generalization. Naive early stopping tends to trap expressive models in local minima typical of Epochs 1--2. To mitigate this: \begin{itemize} \item \textbf{Shock Benchmarks:} We employ a uniform \textbf{3-epoch grace period for all models}. Since convergence rarely occurs within 3 epochs, this ensures models escape initialization artifacts. \item \textbf{Detailed Tables:} We \textbf{do not} employ this grace period in the detailed breakdown tables, preserving raw convergence behaviours for multi-faceted analysis. 
\item \textbf{Smoothing:} While \textit{exponential smoothing} of the validation objective is a another remedy for oscillation that empirically \textit{can} solve this problem, we explicitly \textbf{do not employ it anywhere in this paper} to ensure transparency. \end{itemize}

\subsubsection{Computational and Memory Complexity}
\label{app:lit:complexity}

We summarize the complexity of \Fern{} using the notation from the main text. Let $n$ denote the horizon length, $p$ the patch dimension, and $g$ the number of patches, so that $n = g p$. Let $B$ be the batch size, $d_h$ the hidden width of the MLPs in the coupling blocks, $K_{\text{enc}}$ the number of encoder layers, and $R$ the number of Householder reflections (a hyperparameter, typically $2$–$24$).

\paragraph{Time (FLOPs).}
Each encoder layer operates per patch on vectors $x^i, z^i \in \mathbb{R}^p$ and applies bidirectional affine couplings
\begin{equation*}
\begin{aligned}
h_x^i &= H_x(x^i),          & (s_z^i, t_z^i) &= \phi_x^i(h_x^i), \\
h_z^i &= H_z(z^{i+1}),      & (s_x^i, t_x^i) &= \phi_z^i(h_z^i).
\end{aligned}
\end{equation*}
followed by
\[
z^{i+1} = s_z^i \odot z^i + t_z^i,\qquad x^{i+1} = s_x^i \odot x^i + t_x^i.
\]
Up to constant factors (number of layers per MLP and two heads per layer), one encoder layer costs $O(p d_h)$ per patch, dominated by matrix–vector products. With $K_{\text{enc}}$ layers, the encoder coupling cost over a batch is
\[
O\big(B \cdot g \cdot K_{\text{enc}}\, p d_h\big).
\]

The OT head $\psi(h_z)$ produces $p$ eigenvalues, a shift vector $t_y \in \mathbb{R}^p$, and $R$ Householder vectors in $\mathbb{R}^p$, and applies the resulting SPD map via $R$ reflections. Generation of these parameters via a shallow MLP costs $O(B \cdot g \cdot p d_h)$, and applying the Householder reflections costs $O(B \cdot g \cdot R p)$, so the SPD part adds
\[
O\big(B \cdot g \cdot (p d_h + R p)\big)
\]
FLOPs in the Householder setting. For a dense SPD per patch, the corresponding cost would be $O(B \cdot g \cdot p^2)$ just to apply the map, plus $O(B \cdot g \cdot p^2 d_h)$ if it is predicted by an MLP.

Putting these together, the total time complexity of \Fern{} with Householder SPD is
\begin{align}
\mathcal{T}_{\text{\Fern}}
  &= O\big(B \cdot g \cdot (K_{\text{enc}}\, p d_h + p d_h + R p)\big) \\
  &= O\big(B \cdot g \cdot (K_{\text{enc}}+1)\, p d_h + B \cdot g \cdot R p\big).
\end{align}
and in the dense per-patch SPD setting it would be
\[
O\big(B \cdot g \cdot (K_{\text{enc}}\, p d_h + p^2 d_h + p^2)\big).
\]
In all our experiments we use small fixed constants ($K_{\text{enc}}=5$, $d_h$ fixed, and $R \le p$), so for fixed $g$ and $B$ the head cost scales linearly in the patch dimension $p$ (and hence in the horizon length $n = g p$). For typical settings ($p=24$, $R=4$), the Householder head is roughly $p/R \approx 6\times$ cheaper than a dense $p \times p$ SPD projection.

\paragraph{Memory.}
During training, the dominant memory cost comes from storing activations across encoder layers and patches. We store intermediate $x^i, z^i \in \mathbb{R}^p$ and features $h_x^i, h_z^i$ of width $d_h$ for each of the $K_{\text{enc}}$ layers, yielding
\[
O\big(B \cdot g \cdot K_{\text{enc}} \cdot \max(p, d_h)\big)
\]
activation memory (up to constant factors for the number of such tensors). The spectral parameters add $O(B \cdot g \cdot R p)$ for the Householder vectors and $O(B \cdot g \cdot p)$ for eigenvalues and shifts. Crucially, we never materialize a dense $n \times n$ SPD matrix: the map is kept in a factored Householder–diagonal–Householder form, avoiding the $O(n^2)$ memory footprint of a full covariance.

\subsubsection{Architectural Details}
\label{app:lit:arch}
 
Eigenvalues and translation are parameterized with differentiable soft bounds for numerical stability. We use a soft-clamp that behaves linearly within $[\ell, u]$ and saturates smoothly outside. For the elementwise scaling used in SPD matrix and most bidirectional blocks, we use $s \in [0, 5.5]$ for sensible compromise between expressivity and gradient stability, which is notoriously difficult for this kind of affine coupling structure. We use block-diagonal scaling to mimic complex multiplication---interleaved on the $x$-side in 2 of 5 encoder coupling layers to capture potential rotational dynamics on the data side. For these we use $s \in [-4.5, 4.5]$. Shift vectors $t_y$ use wider bounds $[-15, 15]$.

\subsection{App: Systemss and Datasets} 
\label{app:data}
 
\subsubsection{Chaotic Systems}
Prior work shows that deterministic finite-precision arithmetic can suppress chaos and produce spurious periodic orbits in low precision, and that chaotic systems are highly sensitive to numerical precision and discretization choices. See, e.g., Klöwer et al. (2023) on periodic orbits at low precision and mitigation via stochastic rounding; Teixeira, Reynolds \& Judd (2007) on decoupling times and truncation-error growth; and Liao (2014) on the joint impact of truncation and round-off errors on long-time chaotic simulations. \citep{Klower2023LowPrecChaos,Teixeira2007TimeStep,Liao2014CNS}.
Therefore, all chaotic ODE datasets (Lorenz63, Rössler, Chua, Lorenz96) were generated in \texttt{float64} using the 4th-order Runge-Kutta (RK4) method \cite{butcher1987numerical} and then converted to standard \texttt{float32} Pytorch Tensor. Pilot runs observed materially larger forecast errors across all models with \texttt{float64}.

\paragraph{Lorenz63.}
A canonical 3-D system modeling atmospheric convection:
\[
  \dot x = \sigma(y - x),\quad
  \dot y = x(\rho - z) - y,\quad
  \dot z = xy - \beta z,
\]

\paragraph{Rössler.}
A 3-D system with a folded‐band attractor:
\[
  \dot x = -y - z,\quad
  \dot y = x + a y,\quad
  \dot z = b + z(x - c),
\]

\paragraph{Chua's Circuit}
A 3-D piecewise-linear circuit model with a double-scroll attractor:
\[
\dot x=\alpha\big(y-x-h(x)\big),\quad
\dot y=x-y+z,\quad
\dot z=-\beta\,y,
\]
with
\[
h(x)=m_1 x+\tfrac{1}{2}(m_0-m_1)\big(|x+1|-|x-1|\big).
\]

\paragraph{Lorenz96.}
A $d$-dimensional toy model for mid-latitude atmospheric dynamics, with cyclic nearest-neighbour coupling and constant forcing $\mathrm{forcing}=F$ (parameter \texttt{forcing} in code, dimension \texttt{dim}):
\[
  \dot x_j = \big(x_{j+1} - x_{j-2}\big)\,x_{j-1} - x_j + F,\qquad j=1,\dots,d,
\]
with indices taken modulo $d$ (e.g.\ $x_{0}=x_{d}$, $x_{-1}=x_{d-1}$).

\subsubsection{Stochastic System}
\paragraph{Ornstein--Uhlenbeck (OU).}
A 1-D mean-reverting diffusion with linear drift towards a long-run mean $\mu$ and Gaussian noise, with parameters $(\theta,\mu,\sigma)$ (\texttt{theta}, \texttt{mu}, \texttt{sigma} in code):
\[
  dX_t = \theta\big(\mu - X_t\big)\,dt + \sigma\,dW_t,
\]
integrated with an Euler–Maruyama scheme using step size \texttt{dt}.

\paragraph{Double-well SDE.}
A 1-D bistable diffusion in a double-well potential, parameterized by a shape parameter $a$ and noise scale $\sigma$ (\texttt{a}, \texttt{sigma} in code):
\[
  dX_t = \big(a X_t - X_t^3\big)\,dt + \sigma\,dW_t.
\]
The deterministic drift $aX_t - X_t^3$ creates two metastable wells around $\pm\sqrt{a}$, with noise-driven transitions between wells.

\paragraph{Switching linear (SLDS).}
A 1-D switching linear dynamical system (SLDS) with two linear-Gaussian regimes, specified by $(A_1,Q_1)$ and $(A_2,Q_2)$ and Markov self-transition probabilities $p_{11},p_{22}$ (\texttt{A1}, \texttt{Q1}, \texttt{A2}, \texttt{Q2}, \texttt{p11}, \texttt{p22} in code). Let $s_t\in\{1,2\}$ be the latent regime:
\[
  x_{t+1} = A_{s_t} x_t + \eta_t,\qquad \eta_t \sim \mathcal N(0,Q_{s_t}),
\]
\[
  \mathbb P[s_{t+1}=i \mid s_t=i] = p_{ii},\quad i\in\{1,2\}.
\]
This yields piecewise-linear dynamics with regime switches driven by a 2-state Markov chain.

\paragraph{Seasonal AR.}
A 1-D discrete-time process with an autoregressive term and an explicit seasonal component of period $S$ (\texttt{S}) and slowly drifting amplitude. With AR coefficient $\phi$ (\texttt{phi}), innovation scale $\sigma$ (\texttt{sigma}), initial seasonal amplitude $a_0$ (\texttt{a0}), and linear drift rate \texttt{amp\_drift\_per\_step}, we write
\[
  a_t = a_0 + t\cdot\mathrm{amp\_drift\_per\_step},
\]
\[
  x_t = a_t \cos\!\Bigl(\tfrac{2\pi t}{S}\Bigr) + \phi\,x_{t-1} + \sigma\,\varepsilon_t,\qquad
  \varepsilon_t \sim \mathcal N(0,1).
\]
This models a gradually drifting seasonal pattern on top of an AR(1) background.

\paragraph{GARCH(1,1).}
A discrete-time volatility process with conditionally Gaussian returns and autoregressive conditional variance, parameterized by $(\omega,\alpha,\beta)$ (\texttt{omega}, \texttt{alpha}, \texttt{beta} in code):
\[
  x_t = \sigma_t\,\varepsilon_t,\qquad \varepsilon_t \sim \mathcal N(0,1),
\]
\[
  \sigma_t^2 = \omega + \alpha x_{t-1}^2 + \beta \sigma_{t-1}^2.
\]
Here $x_t$ represents heavy-tailed, volatility-clustered “returns”, while $\sigma_t^2$ evolves as a GARCH(1,1) variance process.



 
\paragraph{Weather}
\label{app:dataset:weather}
21 meteorological indicators at 10 min intervals (Max Planck Institute, Germany, 2020). 

To investigate the air parcel as a complete system, we select the related variables: Selected 14 columns: 'p (mbar)', 'T (degC)', 'Tpot (K)', 'Tdew (degC)', 'rh (\%)', 'VPmax (mbar)', 'VPact (mbar)', 'sh (g/kg)', 'H2OC (mmol/mol)', 'rho (g/m**3)', 'wv (m/s)', 'max. wv (m/s)', 'wd (deg)', 'Tlog (degC)' and eliminate stochastic forcing variables such as rainfall from the model. We apply robust centering and scaling with median and MAD.

\subsection{App: Experiments Setup}
\label{app:exp_setup}
\paragraph{Training setup}
For numerical convenience we optionally use an isotropic base $z \sim \mathcal N(0, a I)$ and $y_{0} \sim \mathcal N(0, a I)$ with a scalar $a > 0$; In both experiments we use $a=0.1$, this simply rescales the eigenvalues in $\Lambda$ (equivalently $\tilde A = \sqrt{a}\, A, \qquad
\Sigma = \tilde A \tilde A^\top.$) and does not change the SPD Jacobian structure or the optimal-transport interpretation. 

We call multiple horizon experiments the \textbf{detailed table} and the main test shock experiments as \textbf{shock table}. Detailed table and shock experiments use different setting (though \textbf{uniformly applied to all models} in respective experiments) for comprehensive and \textbf{varied} angles. We discuss important points:
\begin{itemize}
    \item optimizer: all models are implemented in PyTorch and trained with AdamW~\citep{loshchilov2017decoupled} with no weight decay.
    \item learning rate: shock table uses  $3\times 10^{-4}$, and detailed table uses $9\times 10^{-4}$.
    \item epochs and patience: both $50$ epochs with patience $=5$.
    \item batch size: $95$ for the shock table, and $128$ for main tables.
    \item grace period (validation is logged but cannot trigger early-stopping): $=3$ for shock tables, $=0$ (disabled) for detailed tables.
    \item training objective: \textbf{only} Huber loss ($\delta = 1.0$) for both.
    \item evaluation objective: 0.1·MSE + 1.0·MAE + 0.1·SWD for both, note the SWD difference described right below.
    \item SWD vs WD: (\textit{no projection} for shock tables, $n=500$ projections for detailed tables). No projection reduces SWD to 1D W2 loss so we label `WD' in shock tables.
    \item validating objective smoothing (for early stop): we propose this earlier but \textbf{don't implement it} in any experiments.
    \item input length ($x$-space dim): both tables use \{336\}
    \item forecast horizon ($y$-space dim): shock tables use \{336\}, detailed table use \{96, 192, 336, 720\}.
    \item seeds: shock tables \{7, 1955\}, detailed tables \{7,1955,2023,4\}.
    \item data split: shock tables \textbf{ 70\%/20\%/10\% split} (to make reconstruction and visualization test set easier) and detailed tables \textbf{70\%/10\%/20\% split}. 
    \item pre-processing: only \textbf{zero removal and imputations} enabled for ETTh1 and ETTm1 for shock tables. \textbf{No sentinel value correction for both} experiments.
    \item common practices: all datasets on all experiments are processed without any \texttt{drop\_last} setting
\end{itemize}

\paragraph{Intro to Shock Experiments}

See Table.\ref{tab:app:shock-params} for detailed system parameters. We record the state after every integration step, so the sampling interval of the time series equals the solver step size $dt$. Since train, validation, test split is 0.7, 0.2, 0.1, shock\_frac=0.35 of full data indicates the shock happens at 50\% of the train data. 
 
\begin{table*}[t]
\centering
\scriptsize

\setlength{\tabcolsep}{4pt}
\renewcommand{\arraystretch}{1.1}
\begin{tabular}{llclp{8.3cm}}
\toprule
System & Scenario & $dt$ & steps & Parameters / shock description \\
\midrule
\multicolumn{5}{l}{\textit{Main chaotic benchmarks (no shock).}}\\
Lorenz-63 & main & $0.01$ & $25000$ &
$\texttt{sigma}=10,\ \texttt{rho}=28,\ \texttt{beta}=8/3$. \\[2pt]
Rössler & main & $0.01$ & $25000$ &
$\texttt{a}=0.2,\ \texttt{b}=0.2,\ \texttt{c}=5.7$. \\[2pt]
Chua & main & $0.005$ & $35000$ &
$\texttt{alpha}=15.6,\ \texttt{beta}=28.0,\ \texttt{m0}=-8/7,\ \texttt{m1}=-5/7$. \\
\midrule
\multicolumn{5}{l}{\textit{Synthetic shock scenarios (code identifiers \texttt{PremadeID.*}).}}\\
Lorenz-63 & \texttt{LORENZ\_BASE} & $0.01$ & $35999$ &
Baseline Lorenz-63, no shock; default $\texttt{sigma}=10,\ \texttt{rho}=28,\ \texttt{beta}=8/3$. \\[2pt]
Lorenz-63 & \texttt{LORENZ\_PARAM} & $0.01$ & $35999$ &
Parameter shock (\texttt{shock\_kind = "param"}): 
$\texttt{sigma}: 10 \to 10.1,\ \texttt{rho}: 28 \to 28.1,\ \texttt{beta}: 8/3 \to 8.1/3$. \\[2pt]
Lorenz-63 & \texttt{LORENZ\_STATE} & $0.01$ & $35999$ &
State shock (\texttt{shock\_kind = "state\_eps"}): 
$\texttt{shock\_eps} = 0.9$; ODE parameters as in \texttt{LORENZ\_BASE}. \\[2pt]
Lorenz-63 & \texttt{LORENZ\_SWITCH} & $0.01$ & $35999$ &
Switch shock (\texttt{shock\_kind = "switch"}): 
\texttt{switch\_update} sets $\texttt{rho}: 28 \to 28.1$ and 
$\texttt{initial\_cond}: [1.0, 0.98, 1.1] \to [1.002, 0.982, 1.102]$. \\[2pt]

Rössler & \texttt{ROSSLER\_BASE} & $0.01$ & $35999$ &
Baseline Rössler, no shock; $\texttt{a}=0.2,\ \texttt{b}=0.2,\ \texttt{c}=5.7$. \\[2pt]
Rössler & \texttt{ROSSLER\_PARAM} & $0.01$ & $35999$ &
Parameter shock (\texttt{shock\_kind = "param"}):
$\texttt{a}: 0.2 \to 0.25,\ \texttt{b}: 0.2 \to 0.25,\ \texttt{c}: 5.7 \to 5.75$. \\[2pt]

Lorenz-96 & \texttt{LORENZ96\_BASE} & $0.007$ & $55000$ &
Baseline Lorenz-96; $\texttt{dim} = 6$, $\texttt{forcing} = 8.0$, 
$\texttt{method} = \text{"rk4"}$. \\[2pt]
Lorenz-96 & \texttt{LORENZ96\_SWITCH} & $0.007$ & $55000$ &
Switch shock (\texttt{shock\_kind = "switch"}): 
\texttt{switch\_update} sets $\texttt{forcing}: 8.0 \to 9.0$ and 
$\texttt{initial\_cond} = [0.99, 1.02, 1.02, 1.03, 1.01, 1.01]$ (with $\texttt{dim} = 6$, \texttt{method}=\text{"rk4"}). \\[2pt]

Chua & \texttt{CHUA\_BASE} & $0.005$ & $35999$ &
Baseline Chua, no shock; $\texttt{alpha}=15.6,\ \texttt{beta}=28.0,\ \texttt{m0}=-8/7,\ \texttt{m1}=-5/7$. \\[2pt]
Chua & \texttt{CHUA\_PARAM} & $0.005$ & $35999$ &
Parameter shock (\texttt{shock\_kind = "param"}):
$\texttt{alpha}: 15.6 \to 15.9,\ \texttt{beta}: 28.0 \to 28.5,\ 
\texttt{m0}: -8/7 \to -8.1/7,\ \texttt{m1}: -5/7 \to -5.2/7$. \\[2pt]
Chua & \texttt{CHUA\_SWITCH} & $0.005$ & $35999$ &
Switch shock (\texttt{shock\_kind = "switch"}): 
\texttt{switch\_update} sets 
$\texttt{initial\_cond} = [0.11, 0.01, 0.02]$; other parameters as in \texttt{CHUA\_BASE}. \\[2pt]

OU & \texttt{OU\_BASE} & $0.5$ & $25000$ &
Baseline Ornstein--Uhlenbeck; 
$\texttt{initial\_cond}=[0.0],\ \texttt{theta}=0.2,\ \texttt{mu}=0.0,\ \texttt{sigma}=0.3,\ \texttt{method}$ = \text{"euler"}. \\[2pt]
OU & \texttt{OU\_PARAM} & $0.5$ & $25000$ &
Parameter shock (\texttt{shock\_kind = "param"}): 
$\texttt{mu}: 0.0 \to 0.5$; other OU parameters as in \texttt{OU\_BASE}. \\[2pt]

SLDS & \texttt{SLDS\_BASE} & $0.01$ & $25000$ &
Baseline switching linear dynamical system; 
$\texttt{A1}=0.9,\ \texttt{Q1}=0.05,\ \texttt{A2}=0.98,\ \texttt{Q2}=0.35,\ \texttt{p11}=0.94,\ \texttt{p22}=0.95$. \\[2pt]
SLDS & \texttt{SLDS\_PARAM} & $0.01$ & $25000$ &
Parameter shock (\texttt{shock\_kind = "param"}): 
$\texttt{A1}: 0.9 \to 0.83,\ \texttt{Q1}: 0.05 \to 0.50,\ 
\texttt{A2}: 0.98 \to 0.97,\ \texttt{Q2}: 0.35 \to 0.30,\ 
\texttt{p11}: 0.94 \to 0.96,\ \texttt{p22}: 0.95 \to 0.92$. \\[2pt]
SLDS & \texttt{SLDS\_SWITCH} & $0.01$ & $25000$ &
Switch shock (\texttt{shock\_kind = "switch"}): 
\texttt{switch\_update} sets 
$\texttt{A1}=0.87,\ \texttt{Q1}=0.07,\ \texttt{A2}=0.99,\ \texttt{Q2}=0.45,\ \texttt{p11}=0.90,\ \texttt{p22}=0.95$. \\[2pt]

Double-well & \texttt{DOUBLEWELL\_BASE} & $0.5$ & $25000$ &
Baseline double-well SDE (Euler API, step size $0.5$); 
$\texttt{a}=1.5,\ \texttt{sigma}=0.25,\ \texttt{seed}=1955$. \\[2pt]
Double-well & \texttt{DOUBLEWELL\_PARAM} & $0.5$ & $25000$ &
Parameter shock (\texttt{shock\_kind = "param"}): 
$\texttt{a}: 1.5 \to 1.0,\ \texttt{sigma}: 0.25 \to 0.35$. \\[2pt]
Double-well & \texttt{DOUBLEWELL\_SWITCH} & $0.5$ & $25000$ &
Switch shock (\texttt{shock\_kind = "switch"}): 
\texttt{switch\_update} sets $\texttt{a}=1.0,\ \texttt{sigma}=0.35$ (same target as \texttt{DOUBLEWELL\_PARAM}). \\[2pt]

Seasonal AR & \texttt{SEASONAL\_AR\_BASE} & $0.01$ & $25000$ &
Baseline seasonal AR process (discrete-time; \texttt{dt}/\texttt{method} kept for API): 
$\texttt{S}=24,\ \texttt{phi}=0.5,\ \texttt{sigma}=0.2,\ \texttt{a0}=1.0,\ \texttt{amp\_drift\_per\_step}=0$. \\[2pt]
Seasonal AR & \texttt{SEASONAL\_AR\_PARAM} & $0.01$ & $25000$ &
Parameter shock (\texttt{shock\_kind = "param"}): 
$\texttt{a0}: 1.0 \to 1.4,\ \texttt{sigma}: 0.2 \to 0.35,\ \texttt{phi}: 0.5 \to 0.8$. \\[2pt]

GARCH(1,1) & \texttt{GARCH\_BASE} & $0.01$ & $25000$ &
Baseline GARCH(1,1) volatility model (discrete-time; \texttt{dt}/\texttt{method} kept for API): 
$\texttt{omega}=0.01,\ \texttt{alpha}=0.06,\ \texttt{beta}=0.90$. \\[2pt]
GARCH(1,1) & \texttt{GARCH\_PARAM} & $0.01$ & $25000$ &
Parameter shock (\texttt{shock\_kind = "param"}): 
$\texttt{omega}: 0.01 \to 0.03,\ \texttt{alpha}: 0.06 \to 0.15,\ \texttt{beta}: 0.90 \to 0.70$. \\[2pt]

KS & \texttt{KS\_BASE} & $0.01$ & $25000$ &
Baseline Kuramoto--Sivashinsky; 
$\texttt{nx}=64,\ \texttt{Lx}=22.0,\ \texttt{nu}=1.0,\ \texttt{method}$ = \text{"etdrk4"}. \\[2pt]
KS & \texttt{KS\_PARAM} & $0.01$ & $25000$ &
Parameter shock (\texttt{shock\_kind = "param"}): 
$\texttt{nu}: 1.0 \to 0.80$ (other KS parameters as in \texttt{KS\_BASE}). \\
\bottomrule
\end{tabular}
\caption{Parameter settings for the main chaotic benchmarks (top block) and all synthetic shock scenarios used in our experiments (bottom block). Shock scenarios are instantiated via \texttt{PremadeID.*} in \texttt{make\_source}, with \texttt{shock\_frac} fixed to $0.35$ so that shocks are applied after the first 35\% of the trajectory.}
\label{tab:app:shock-params}
\end{table*}

\clearpage
\subsection{App: Ablation, Footprint and Detail Tables}
\label{sec:ablation_summary_main_reduced} 

\begin{table*}[t]
\centering
\small
\setlength{\tabcolsep}{3pt}
\begin{tabular}{p{5.8cm}rrrrr}
\toprule
\textbf{Variant (dataset)} & \textbf{MSE↓} & \textbf{MAE↓} & \textbf{WD↓} & \textbf{EPT↑} & \textbf{Time (s)} \\
\midrule 
Base (ETTh1)   & 10.96 & 1.88 &  5.75 &  63.34 & --- \\
Base (Lorenz63)& \textbf{21.66} & \textbf{2.39} & \textbf{4.41} & 241.25 & --- \\
\midrule 
No encoder \& no mean updates (Lorenz63) 
  & 194.43 & 10.99 & 189.47 & 17.10 & 492.73 \\
\midrule 
Only encoder (ETTh1) 
  & 11.17 & 1.87 &  5.86 &  63.00 & 378.93 \\
Only encoder (Lorenz63) 
  & 27.09 & 2.85 &  6.17 & 214.10 & 1127.98 \\
\midrule 
No rotation (ETTh1) 
  & 11.84 & 1.90 &  7.72 &  66.00 & 617.55 \\
No rotation (Lorenz63) 
  & 27.62 & 2.94 &  6.19 & 203.60 & 846.38 \\
\midrule 
No patching (ETTh1) 
  & 10.99 & \textbf{1.84} &  5.38 &  64.10 & 422.65 \\
No patching (Lorenz63) 
  & 22.86 & 2.50 &  4.47 & 231.10 & 1239.11 \\
\midrule 
Reflection = 2 (ETTh1) 
  & 11.52 & 1.89 &  6.02 &  64.30 & 429.67 \\
Reflection = 2 (Lorenz63) 
  & 26.14 & 2.79 &  6.51 & 199.10 & 889.23 \\
\midrule 
Reflection = 24 (8-block) (ETTh1) 
  & \textbf{10.91} & 1.87 & \textbf{5.21} &  63.00 & 432.86 \\
Reflection = 24 (8-block) (Lorenz63) 
  & 23.92 & 2.70 &  5.70 & 213.00 & 930.42 \\
\bottomrule
\end{tabular}
\caption{Ablations of FERN components at prediction length 192 on ETTh1, and Lorenz-63. Base uses reflection=8 and patch size P=24; \textbf{Bold} marks the best (lowest) MSE, MAE, WD.}
\label{tab:app:ablations-192-stacked}
\end{table*}

\begin{table*}[t]
\centering
\setlength{\tabcolsep}{1.5pt}
\footnotesize
\begin{tabular}{l*{12}{r}}
\toprule
&
  \multicolumn{3}{c}{fr} &
  \multicolumn{3}{c}{tm} &
  \multicolumn{3}{c}{tst} &
  \multicolumn{3}{c}{dl} \\
\cmidrule(lr){2-4}\cmidrule(lr){5-7}\cmidrule(lr){8-10}\cmidrule(lr){11-13}
Data &
  MSE & MAE & WD &
  MSE & MAE & WD &
  MSE & MAE & WD &
  MSE & MAE & WD \\
\midrule
Lorenz
  & \first{21.82}{\tiny $\pm$ 2.13} & \first{2.17} & \first{2.23}
  & 30.94{\tiny $\pm$ 5.62}        & \second{3.19} & 11.11
  & \second{30.11}{\tiny $\pm$ 2.92} & 3.28        & \second{9.60}
  & 67.76{\tiny $\pm$ 1.12}        & 6.07         & 38.22 \\
\addlinespace[1pt]
Rossler
  & \first{0.04}{\tiny $\pm$ 0.01} & \first{0.11} & \first{0.02}
  & \second{6.01}{\tiny $\pm$ 0.26} & \second{1.09} & \second{5.20}
  & 8.33{\tiny $\pm$ 0.36}         & 1.43         & 7.25
  & 11.64{\tiny $\pm$ 0.45}        & 1.82         & 10.20 \\
\addlinespace[1pt]
Chua
  & \first{0.08}{\tiny $\pm$ 0.13} & \first{0.08} & \first{0.05}
  & \second{0.20}{\tiny $\pm$ 0.21} & \second{0.16} & \second{0.15}
  & 0.49{\tiny $\pm$ 0.13}         & 0.32         & 0.37
  & 0.39{\tiny $\pm$ 0.02}         & 0.30         & 0.24 \\
\addlinespace[1pt] 
ETTh1
  & \first{6.60}{\tiny $\pm$ 0.11}  & \second{1.53} & \first{2.64}
  & 6.83{\tiny $\pm$ 0.16}         & \first{1.52}  & 2.83
  & \second{6.62}{\tiny $\pm$ 0.14} & 1.54         & 2.77
  & 7.04{\tiny $\pm$ 0.06}         & \second{1.53} & \second{2.75} \\
\addlinespace[1pt]
ETTm1
  & 5.80{\tiny $\pm$ 0.25}         & 1.45         & 2.85
  & \first{5.27}{\tiny $\pm$ 0.22} & \second{1.39} & \second{2.60}
  & \second{5.36}{\tiny $\pm$ 0.42} & \first{1.37} & \first{2.44}
  & 6.31{\tiny $\pm$ 0.11}         & \second{1.39} & 3.65 \\

\bottomrule
\end{tabular}

\vspace{2pt}
\caption{Aggregated errors across prediction horizons $H \in \{96,192,336,720\}$ with input length=$336$. Baselines: \textbf{fr} = FERN, \textbf{tm} = TimeMixer, \textbf{tst} = PatchTST, \textbf{dl} = DLinear. Values are means over 4 seeds [7, 1955, 2023, 4]; tiny ± indicates standard error computed from per-horizon standard errors (see text). *Weather is normalized. Best and second-best values across models for each row/metric are highlighted via \first and \second.}
\label{tab:main_results}
\end{table*}

\paragraph{Short Note}
Compute footprint tables and detailed ablation table is place along side the detailed tables. Training logs are retained for transparency and available upon request. detailed tables and shock tables use different protocols, see~\ref{app:exp_setup}. Each table is internally consistent within its regime; We therefore avoid direct quantitative comparisons across regimes unless we explicitly re-run models under a matched setup.

\begin{table}[t]
\centering
\small
\setlength{\tabcolsep}{3pt}
\begin{tabular}{lcccc}
\toprule
\textbf{Model} & \textbf{Dataset} & \textbf{Training time} & \textbf{Params (M)} & \textbf{GFLOPs (G)} \\
\midrule
\Fern     & L63 & 16.0 & 1.025 & 0.0035 \\
\Fern     & m1  & 83   & 1.025 & 0.0035 \\
TimeMixer & L63 & 21.0 & 0.886 & 0.0463 \\
TimeMixer & m1  & 120  & 0.886 & 0.0463 \\
PatchTST  & L63 & 10.0 & 2.008 & 0.0308 \\
PatchTST  & m1  & 110  & 2.008 & 0.0308 \\
DLinear   & L63 & 2.5  & 0.679 & 0.0007 \\
DLinear   & m1  & 27   & 0.679 & 0.0007 \\
\bottomrule
\end{tabular}
\caption{Compute footprint on 52k steps for Lorenz63 (336-in-336-out) in minutes and ETTm1 (96-in-336-out) in seconds. GFLOPs are reported per sample per step.}
\label{app:tab:compute_time}
\end{table}

\begin{table*}[t]
\centering
\setlength{\tabcolsep}{3pt}
\begin{tabular}{lllrrrr}
\toprule
Data & Hor. & Model & \multicolumn{1}{c}{MSE} & \multicolumn{1}{c}{MAE} & \multicolumn{1}{c}{SWD} & \multicolumn{1}{c}{EPT} \\
\midrule
\multirow{16}{*}{Chua}
& \multirow{4}{*}{96}  & \Fern      & 0.0007 \( \pm \) 0.00 & 0.0191 \( \pm \) 0.01 & 0.0007 \( \pm \) 0.00 & 96 \\
&                       & TimeMixer  & 0.0015 \(\pm\) 0.00   & 0.0259 \(\pm\) 0.00   & 0.0014 \(\pm\) 0.00   & 96 \\
&                       & PatchTST   & 0.0063 \(\pm\) 0.00   & 0.0509 \(\pm\) 0.01   & 0.0057 \(\pm\) 0.00   & 96 \\
&                       & DLinear    & 0.0168 \(\pm\) 0.00   & 0.0796 \(\pm\) 0.01   & 0.0167 \(\pm\) 0.00   & 96 \\
\cmidrule(lr){2-7}
& \multirow{4}{*}{192} & \Fern      & 0.0011 \( \pm \) 0.00 & 0.0257 \( \pm \) 0.00 & 0.0009 \( \pm \) 0.00 & 192 \\
&                       & TimeMixer  & 0.0349 \(\pm\) 0.01   & 0.1199 \(\pm\) 0.01   & 0.0319 \(\pm\) 0.01   & 192 \\
&                       & PatchTST   & 0.1242 \(\pm\) 0.06   & 0.1924 \(\pm\) 0.05   & 0.1091 \(\pm\) 0.05   & 192 \\
&                       & DLinear    & 0.0265 \(\pm\) 0.01   & 0.1183 \(\pm\) 0.02   & 0.0246 \(\pm\) 0.01   & 192 \\
\cmidrule(lr){2-7}
& \multirow{4}{*}{336} & \Fern      & 0.0010 \( \pm \) 0.00 & 0.0225 \( \pm \) 0.00 & 0.0004 \( \pm \) 0.00 & 336 \\
&                       & TimeMixer  & 0.0045 \(\pm\) 0.00   & 0.0479 \(\pm\) 0.01   & 0.0028 \(\pm\) 0.00   & 336 \\
&                       & PatchTST   & 0.0278 \(\pm\) 0.01   & 0.1060 \(\pm\) 0.02   & 0.0215 \(\pm\) 0.01   & 336 \\
&                       & DLinear    & 0.1590 \(\pm\) 0.01   & 0.2703 \(\pm\) 0.01   & 0.0813 \(\pm\) 0.01   & 336 \\
\cmidrule(lr){2-7}
& \multirow{4}{*}{720} & \Fern      & 0.3301 \( \pm \) 0.51 & 0.2386 \( \pm \) 0.17 & 0.1994 \( \pm \) 0.33 & 474 \( \pm \) 73 \\
&                       & TimeMixer  & 0.7624 \(\pm\) 0.82   & 0.4475 \(\pm\) 0.26   & 0.5685 \(\pm\) 0.64   & 427 \(\pm\) 80 \\
&                       & PatchTST   & 1.7924 \(\pm\) 0.52   & 0.9209 \(\pm\) 0.11   & 1.3499 \(\pm\) 0.43   & 228 \(\pm\) 45 \\
&                       & DLinear    & 1.3719 \(\pm\) 0.09   & 0.7314 \(\pm\) 0.03   & 0.8305 \(\pm\) 0.07   & 119 \(\pm\) 2 \\
\addlinespace[2pt]
\midrule
\multicolumn{2}{r}{\textit{Avg}} & \Fern      & 0.0832  & 0.0765  & 0.0504  & 274.4 \\
\multicolumn{2}{r}{}            & TimeMixer  & 0.20083 & 0.16030 & 0.15115 & 262.7 \\
\multicolumn{2}{r}{}            & PatchTST   & 0.48767 & 0.31755 & 0.37155 & 213.0 \\
\multicolumn{2}{r}{}            & DLinear    & 0.39355 & 0.29990 & 0.23828 & 185.6 \\
\addlinespace[2pt]
\bottomrule
\end{tabular}
\caption{Chua's circuit, dt=5e-3, step=35,000. Values are mean \( \pm \) s.e. across 4 seeds. Higher is better for EPT, lower is better for the rest.}
\label{app:table:chua}
\end{table*}

\begin{table*}[t]

\centering
\setlength{\tabcolsep}{3pt}
\begin{tabular}{lllrrrr}
\toprule
Data & Hor. & Model & \multicolumn{1}{c}{MSE} & \multicolumn{1}{c}{MAE} & \multicolumn{1}{c}{SWD} & \multicolumn{1}{c}{EPT} \\
\midrule
\multirow{16}{*}{Rossler}
& \multirow{4}{*}{96}  & \Fern      & 0.0032 \( \pm \) 0.00 & 0.0445 \( \pm \) 0.01 & 0.0030 \( \pm \) 0.00 & 96 \\ 
&                       & TimeMixer  & 0.1033 \(\pm\) 0.07   & 0.2454 \(\pm\) 0.09   & 0.0963 \(\pm\) 0.07   & 96 \\
&                       & PatchTST   & 0.1846 \(\pm\) 0.10   & 0.3155 \(\pm\) 0.09   & 0.1700 \(\pm\) 0.10   & 96 \\
&                       & DLinear    & 0.3036 \(\pm\) 0.09   & 0.3343 \(\pm\) 0.06   & 0.2908 \(\pm\) 0.08   & 96 \\
\cmidrule(lr){2-7}
& \multirow{4}{*}{192} & \Fern      & 0.0066 \( \pm \) 0.00 & 0.0618 \( \pm \) 0.01 & 0.0055 \( \pm \) 0.00 & 192 \\ 
&                       & TimeMixer  & 0.1124 \(\pm\) 0.03   & 0.2689 \(\pm\) 0.03   & 0.1016 \(\pm\) 0.03   & 192 \\
&                       & PatchTST   & 0.5704 \(\pm\) 0.14   & 0.5893 \(\pm\) 0.09   & 0.4762 \(\pm\) 0.14   & 192 \\
&                       & DLinear    & 3.6638 \(\pm\) 1.14   & 1.0719 \(\pm\) 0.19   & 3.2148 \(\pm\) 1.11   & 179 \(\pm\) 8 \\
\cmidrule(lr){2-7}
& \multirow{4}{*}{336} & \Fern      & 0.0830 \( \pm \) 0.03 & 0.1988 \( \pm \) 0.04 & 0.0567 \( \pm \) 0.02 & 332 \( \pm \) 4 \\ 
&                       & TimeMixer  & 14.9881 \(\pm\) 0.84  & 2.4916 \(\pm\) 0.09   & 12.8596 \(\pm\) 0.74  & 162 \(\pm\) 3 \\
&                       & PatchTST   & 23.9804 \(\pm\) 1.43  & 3.3559 \(\pm\) 0.12   & 20.8028 \(\pm\) 1.51  & 69 \(\pm\) 1 \\
&                       & DLinear    & 34.3263 \(\pm\) 1.34  & 4.3221 \(\pm\) 0.09   & 29.8861 \(\pm\) 1.22  & 60 \(\pm\) 1 \\
\cmidrule(lr){2-7}
& \multirow{4}{*}{720} & \Fern      & 0.0548 \( \pm \) 0.04 & 0.1258 \( \pm \) 0.03 & 0.0160 \( \pm \) 0.01 & 677 \( \pm \) 44 \\ 
&                       & TimeMixer  & 8.8210 \(\pm\) 0.60   & 1.3566 \(\pm\) 0.06   & 7.7532 \(\pm\) 0.47   & 557 \(\pm\) 21 \\
&                       & PatchTST   & 8.5990 \(\pm\) 0.19   & 1.4545 \(\pm\) 0.02   & 7.5545 \(\pm\) 0.16   & 539 \(\pm\) 3 \\
&                       & DLinear    & 8.2653 \(\pm\) 0.34   & 1.5395 \(\pm\) 0.09   & 7.3939 \(\pm\) 0.29   & 594 \(\pm\) 24 \\
\addlinespace[2pt]
\midrule
\multicolumn{2}{r}{\textit{Avg}} & \Fern      & 0.0369 & 0.1077 & 0.0203 & 324.2 \\
\multicolumn{2}{r}{}                       & TimeMixer  & 6.0062 & 1.0906 & 5.2027 & 251.8 \\
\multicolumn{2}{r}{}                       & PatchTST   & 8.3336 & 1.4288 & 7.2509 & 223.9 \\
\multicolumn{2}{r}{}                       & DLinear    & 11.6398 & 1.8170 & 10.1964 & 232.4 \\
\bottomrule
\end{tabular}
\caption{Rossler, dt=1e-2, step=25,000. Values are mean \( \pm \) s.e. across 4 seeds. Higher is better for EPT, lower is better for the rest.}
\label{app:table:rossler}
\end{table*}

\begin{table*}[t]

\centering
\setlength{\tabcolsep}{3pt}
\begin{tabular}{lllrrrr}
\toprule
Data & Hor. & Model & \multicolumn{1}{c}{MSE} & \multicolumn{1}{c}{MAE} & \multicolumn{1}{c}{SWD} & \multicolumn{1}{c}{EPT} \\
\midrule
\multirow{16}{*}{ETTh1}
& \multirow{4}{*}{96}  & \Fern      & 6.68 \( \pm \) 0.05 & 1.50 \( \pm \) 0.01 & 2.88 \( \pm \) 0.09 & 34.11 \( \pm \) 0.68 \\ 
&                       & TimeMixer  & 6.85 \(\pm\) 0.43   & 1.47 \(\pm\) 0.04   & 3.02 \(\pm\) 0.28   & 39.40 \(\pm\) 1.31 \\
&                       & PatchTST   & 5.87 \(\pm\) 0.14   & 1.43 \(\pm\) 0.03   & 2.48 \(\pm\) 0.08   & 41.45 \(\pm\) 1.15 \\
&                       & DLinear    & 6.62 \(\pm\) 0.06   & 1.45 \(\pm\) 0.01   & 2.77 \(\pm\) 0.10   & 38.53 \(\pm\) 1.15 \\
\cmidrule(lr){2-7}
& \multirow{4}{*}{192} & \Fern      & 6.21 \( \pm \) 0.19 & 1.49 \( \pm \) 0.03 & 1.81 \( \pm \) 0.13 & 61.29 \( \pm \) 7.42 \\ 
&                       & TimeMixer  & 6.82 \(\pm\) 0.08   & 1.49 \(\pm\) 0.02   & 2.21 \(\pm\) 0.10   & 70.84 \(\pm\) 4.58 \\
&                       & PatchTST   & 6.12 \(\pm\) 0.24   & 1.48 \(\pm\) 0.03   & 1.81 \(\pm\) 0.11   & 78.49 \(\pm\) 8.21 \\
&                       & DLinear    & 7.39 \(\pm\) 0.17   & 1.51 \(\pm\) 0.03   & 2.36 \(\pm\) 0.17   & 81.34 \(\pm\) 6.13 \\
\cmidrule(lr){2-7}
& \multirow{4}{*}{336} & \Fern      & 6.41 \( \pm \) 0.33 & 1.54 \( \pm \) 0.03 & 3.25 \( \pm \) 0.34 & 68.77 \( \pm \) 7.36 \\ 
&                       & TimeMixer  & 6.06 \(\pm\) 0.20   & 1.43 \(\pm\) 0.01   & 3.27 \(\pm\) 0.28   & 72.88 \(\pm\) 1.20 \\
&                       & PatchTST   & 6.81 \(\pm\) 0.47   & 1.56 \(\pm\) 0.05   & 3.75 \(\pm\) 0.63   & 69.49 \(\pm\) 1.49 \\
&                       & DLinear    & 6.39 \(\pm\) 0.05   & 1.48 \(\pm\) 0.01   & 3.16 \(\pm\) 0.11   & 74.75 \(\pm\) 2.14 \\
\cmidrule(lr){2-7}
& \multirow{4}{*}{720} & \Fern      & 7.09 \( \pm \) 0.20 & 1.57 \( \pm \) 0.06 & 2.60 \( \pm \) 0.19 & 112.16 \( \pm \) 18.48 \\ 
&                       & TimeMixer  & 7.57 \(\pm\) 0.45   & 1.69 \(\pm\) 0.07   & 2.80 \(\pm\) 0.35   & 125.88 \(\pm\) 2.06 \\
&                       & PatchTST   & 7.67 \(\pm\) 0.14   & 1.70 \(\pm\) 0.05   & 3.03 \(\pm\) 0.15   & 121.38 \(\pm\) 1.47 \\
&                       & DLinear    & 7.77 \(\pm\) 0.15   & 1.67 \(\pm\) 0.05   & 2.70 \(\pm\) 0.11   & 124.28 \(\pm\) 0.78 \\
\addlinespace[2pt]
\midrule
\multicolumn{2}{r}{\textit{Simple Average}}  & \Fern      & 6.60 & 1.53 & 2.64 & 69.08 \\
\multicolumn{2}{r}{}                        & TimeMixer  & 6.83 & 1.52 & 2.83 & 77.25 \\
\multicolumn{2}{r}{}                        & PatchTST   & 6.62 & 1.54 & 2.77 & 77.70 \\
\multicolumn{2}{r}{}                        & DLinear    & 7.04 & 1.53 & 2.75 & 79.72 \\
\bottomrule
\end{tabular}

\caption{ETTh1. Values are mean \( \pm \) s.e. across 4 seeds. Higher is better for EPT, lower is better for the rest.}
\label{app:table:etth1}
\end{table*}

\begin{table*}[t]

\centering
\setlength{\tabcolsep}{3pt}
\begin{tabular}{lllrrrr}
\toprule
Data & Hor. & Model & \multicolumn{1}{c}{MSE} & \multicolumn{1}{c}{MAE} & \multicolumn{1}{c}{SWD} & \multicolumn{1}{c}{EPT} \\
\midrule
\multirow{16}{*}{ETTm1}
& \multirow{4}{*}{96}  & \Fern      & 2.67 \(\pm\) 0.03 & 1.07 \(\pm\) 0.03 & 0.95 \(\pm\) 0.08 & 44.32 \(\pm\) 1.27 \\
&                       & TimeMixer  & 2.91 \(\pm\) 0.64 & 0.98 \(\pm\) 0.09 & 1.07 \(\pm\) 0.45 & 46.79 \(\pm\) 2.80 \\
&                       & PatchTST   & 3.03 \(\pm\) 0.36 & 1.04 \(\pm\) 0.06 & 1.14 \(\pm\) 0.31 & 43.57 \(\pm\) 0.35 \\
&                       & DLinear    & 2.33 \(\pm\) 0.11 & 0.91 \(\pm\) 0.03 & 1.01 \(\pm\) 0.11 & 43.32 \(\pm\) 2.22 \\
\cmidrule(lr){2-7}
& \multirow{4}{*}{192} & \Fern      & 6.86 \(\pm\) 0.69 & 1.59 \(\pm\) 0.06 & 5.30 \(\pm\) 0.58 & 66.75 \(\pm\) 1.64 \\
&                       & TimeMixer  & 5.77 \(\pm\) 0.48 & 1.62 \(\pm\) 0.09 & 4.32 \(\pm\) 0.38 & 69.97 \(\pm\) 4.67 \\
&                       & PatchTST   & 6.60 \(\pm\) 1.60 & 1.54 \(\pm\) 0.13 & 4.96 \(\pm\) 1.45 & 67.17 \(\pm\) 2.99 \\
&                       & DLinear    & 7.10 \(\pm\) 0.33 & 1.50 \(\pm\) 0.04 & 5.60 \(\pm\) 0.36 & 69.99 \(\pm\) 1.90 \\
\cmidrule(lr){2-7}
& \multirow{4}{*}{336} & \Fern      & 6.42 \(\pm\) 0.56 & 1.57 \(\pm\) 0.05 & 3.45 \(\pm\) 0.72 & 105.69 \(\pm\) 4.66 \\
&                       & TimeMixer  & 6.04 \(\pm\) 0.32 & 1.55 \(\pm\) 0.08 & 3.61 \(\pm\) 0.41 & 107.72 \(\pm\) 3.85 \\
&                       & PatchTST   & 5.75 \(\pm\) 0.22 & 1.45 \(\pm\) 0.06 & 2.31 \(\pm\) 0.50 & 101.94 \(\pm\) 4.42 \\
&                       & DLinear    & 7.98 \(\pm\) 0.23 & 1.56 \(\pm\) 0.04 & 5.37 \(\pm\) 0.28 & 111.09 \(\pm\) 0.47 \\
\cmidrule(lr){2-7}
& \multirow{4}{*}{720} & \Fern      & 7.26 \(\pm\) 0.43 & 1.58 \(\pm\) 0.05 & 1.68 \(\pm\) 0.18 & 162.74 \(\pm\) 16.46 \\
&                       & TimeMixer  & 6.35 \(\pm\) 0.15 & 1.41 \(\pm\) 0.03 & 1.39 \(\pm\) 0.28 & 207.69 \(\pm\) 15.89 \\
&                       & PatchTST   & 6.06 \(\pm\) 0.14 & 1.44 \(\pm\) 0.02 & 1.35 \(\pm\) 0.22 & 212.47 \(\pm\) 9.95 \\
&                       & DLinear    & 7.82 \(\pm\) 0.07 & 1.58 \(\pm\) 0.00 & 2.61 \(\pm\) 0.15 & 186.79 \(\pm\) 13.50 \\
\addlinespace[2pt]
\midrule
\multicolumn{2}{r}{\textit{Avg}} & \Fern     & 5.80 & 1.45 & 2.85 & 94.88 \\
\multicolumn{2}{r}{}            & TimeMixer & 5.27 & 1.39 & 2.60 & 108.04 \\
\multicolumn{2}{r}{}            & PatchTST  & 5.36 & 1.37 & 2.44 & 106.29 \\
\multicolumn{2}{r}{}            & DLinear   & 6.31 & 1.39 & 3.65 & 102.80 \\
\bottomrule
\end{tabular}

\caption{ETTm1. Values are mean \( \pm \) s.e. across 4 seeds. Higher is better for EPT, lower is better for the rest.}
\label{app:table:ettm1-upd}
\end{table*}

\begin{table*}[t]

\centering
\setlength{\tabcolsep}{3pt}
\begin{tabular}{lllrrrr}
\toprule
Data & Hor. & Model & \multicolumn{1}{c}{MSE} & \multicolumn{1}{c}{MAE} & \multicolumn{1}{c}{SWD} & \multicolumn{1}{c}{EPT} \\
\midrule
\multirow{16}{*}{Lorenz}
& \multirow{4}{*}{96}  & \Fern      & 0.47 \( \pm \) 0.15 & 0.50 \( \pm \) 0.08 & 0.21 \( \pm \) 0.06 & 96.00 \( \pm \) 0.00 \\ 
&                       & TimeMixer  & 0.18 \(\pm\) 0.01   & 0.33 \(\pm\) 0.01   & 0.10 \(\pm\) 0.01   & 96.00 \(\pm\) 0.00 \\
&                       & PatchTST   & 1.17 \(\pm\) 0.31   & 0.83 \(\pm\) 0.09   & 0.86 \(\pm\) 0.22   & 96.00 \(\pm\) 0.00 \\
&                       & DLinear    & 54.04 \(\pm\) 3.03  & 4.92 \(\pm\) 0.17   & 33.48 \(\pm\) 1.79  & 50.42 \(\pm\) 2.88 \\
\cmidrule(lr){2-7}
& \multirow{4}{*}{192} & \Fern      & 2.06 \( \pm \) 0.69 & 0.83 \( \pm \) 0.13 & 0.33 \( \pm \) 0.09 & 175.23 \( \pm \) 6.90 \\ 
&                       & TimeMixer  & 16.91 \(\pm\) 16.46 & 2.45 \(\pm\) 1.20   & 8.55 \(\pm\) 8.93   & 136.16 \(\pm\) 56.75 \\
&                       & PatchTST   & 6.90 \(\pm\) 2.06   & 1.91 \(\pm\) 0.29   & 2.52 \(\pm\) 0.80   & 179.67 \(\pm\) 12.43 \\
&                       & DLinear    & 79.25 \(\pm\) 2.61  & 6.78 \(\pm\) 0.15   & 48.51 \(\pm\) 1.24  & 19.76 \(\pm\) 0.18 \\
\cmidrule(lr){2-7}
& \multirow{4}{*}{336} & \Fern      & 21.25 \( \pm \) 8.18 & 2.39 \( \pm \) 0.41 & 3.49 \( \pm \) 1.38 & 187.03 \( \pm \) 12.57 \\ 
&                       & TimeMixer  & 55.97 \(\pm\) 2.09   & 5.10 \(\pm\) 0.07   & 25.41 \(\pm\) 1.99  & 92.78 \(\pm\) 7.91 \\
&                       & PatchTST   & 60.96 \(\pm\) 9.40   & 5.43 \(\pm\) 0.44   & 25.12 \(\pm\) 2.81  & 105.74 \(\pm\) 14.05 \\
&                       & DLinear    & 61.06 \(\pm\) 1.31   & 5.64 \(\pm\) 0.12   & 30.28 \(\pm\) 0.37  & 79.80 \(\pm\) 1.02 \\
\cmidrule(lr){2-7}
& \multirow{4}{*}{720} & \Fern      & 63.52 \( \pm \) 2.25 & 4.96 \( \pm \) 0.20 & 4.89 \( \pm \) 0.57 & 293.46 \( \pm \) 14.90 \\ 
&                       & TimeMixer  & 50.71 \(\pm\) 15.19  & 4.91 \(\pm\) 0.40   & 10.39 \(\pm\) 3.56  & 247.14 \(\pm\) 13.47 \\
&                       & PatchTST   & 51.41 \(\pm\) 6.62   & 4.94 \(\pm\) 0.24   & 9.89 \(\pm\) 2.96   & 254.76 \(\pm\) 29.58 \\
&                       & DLinear    & 76.69 \(\pm\) 1.48   & 6.96 \(\pm\) 0.11   & 40.59 \(\pm\) 0.29  & 24.01 \(\pm\) 3.50 \\
\addlinespace[2pt]
\midrule
\multicolumn{2}{r}{\textit{Avg}} & \Fern     & 21.82 & 2.17 & 2.23 & 187.93 \\
\multicolumn{2}{r}{}            & TimeMixer & 30.94 & 3.19 & 11.11 & 143.02 \\
\multicolumn{2}{r}{}            & PatchTST  & 30.11 & 3.28 & 9.60 & 159.04 \\
\multicolumn{2}{r}{}            & DLinear   & 67.76 & 6.07 & 38.22 & 43.50 \\
\bottomrule
\end{tabular}

\caption{Lorenz. Values are mean \( \pm \) s.e. across 4 seeds. Higher is better for EPT, lower is better for the rest.}
\label{app:table:lorenz}
\end{table*}

\clearpage

\end{document}